\documentclass[manuscript]{acmart}

\AtBeginDocument{%
  }

\usepackage{booktabs}
\usepackage{pdflscape}
\usepackage{colortbl}
\usepackage{svg}
\usepackage{ragged2e}
\usepackage{array}
\usepackage{amsmath}

\usepackage{longtable}
\usepackage{multirow}
\usepackage{tabularx}

\usepackage{subcaption}

\usepackage{graphicx}
\usepackage[export]{adjustbox}
\usepackage{float}
\usepackage{wrapfig}

\begin{document}


\title{Critical Challenges and Guidelines in Evaluating Synthetic Tabular Data: A Systematic Review}


\author{Nazia Nafis}
\affiliation{%
  \institution{Healthy Lifespan Institute, School of Computer Science, University of Sheffield}
  \city{Sheffield}
  \country{UK}
}
\email{nnafis1@sheffield.ac.uk}

\author{Iñaki Esnaola}
\affiliation{%
  \institution{School of Electrical and Electronic Engineering, University of Sheffield}
  \city{Sheffield}
  \country{UK}
}
\email{esnaola@sheffield.ac.uk}

\author{Alvaro Martinez-Perez}
\affiliation{%
  \institution{Healthy Lifespan Institute, School of Sociological Studies, Politics and International Relations, University of Sheffield}
  \city{Sheffield}
  \country{UK}
}
\email{a.martinez-perez@sheffield.ac.uk}

\author{Maria-Cruz Villa-Uriol}
\affiliation{%
  \institution{Healthy Lifespan Institute, School of Computer Science, University of Sheffield}
  \city{Sheffield}
  \country{UK}
}
\email{m.villa-uriol@sheffield.ac.uk}

\author{Venet Osmani}
\affiliation{%
  \institution{Digital Environment Research Institute, Queen Mary University of London}
  \city{London}
  \country{UK}
}
\email{v.osmani@qmul.ac.uk}


\renewcommand{\shortauthors}{Nafis et al.}

\begin{abstract}
\textbf{Abstract.} Generating synthetic tabular health data is challenging, and evaluating their quality is equally, if not more, complex. This systematic review highlights the critical importance of rigorous evaluation of synthetic health data to ensure reliability, clinical relevance, and appropriate use. From an initial identification of $2067$ relevant papers published in the last ten years, $134$ studies were selected for detailed analysis. Our review identifies key challenges, including lack of consensus on evaluation methods, inconsistent application of evaluation metrics, limited involvement of domain experts, inadequate reporting of dataset characteristics, and limited reproducibility of results. In response, we provide a structured consolidation of synthetic data generation and evaluation methods into taxonomies, alongside practical guidelines to support more robust and standardised evaluation practices. These findings aim to support the responsible development and use of synthetic health data, aligned with emerging expectations around transparency, reproducibility, and governance, ultimately enabling the community to fully harness its transformative potential and accelerate innovation.
\end{abstract}

\ccsdesc[500]{Computing methodologies~Machine learning approaches}
\ccsdesc[500]{Applied computing~Health informatics}

\keywords{synthetic data, tabular data, time series data}


\maketitle

\section{Introduction}

Access to high-quality data is fundamental to advancing scientific research. In disciplines such as healthcare, data is pivotal to enhance patient care, optimise resource management, and enable the discovery of new medical insights, particularly with the rise of artificial intelligence. Structured health data, such as tabular electronic health records, have been recognised as having one of the highest potential to provide timely and relevant information in clinical decision-making \cite{Tayefi2021ChallengesAO}. However, complex data-sharing governance rules have resulted in health data being locked away in isolated silos \cite{articleCruz,unknownAsiimwe}, where they generally remain inaccessible except to a few researchers \cite{inproceedingsLefebvre}. This inevitably hampers reproducible health research, hindering the advancement of patient care and impeding the future potential of clinical artificial intelligence \cite{articleBernardi}.

There is an urgent need to democratise access to health data \cite{inproceedingsLefebvre} without losing sight of patient privacy and confidentiality \cite{articleSmit}. In this context, synthetic health data emerges as an attractive solution to address this challenge, and institutions worldwide are increasingly recognising their potential, while also raising new challenges around data fidelity, bias propagation, and privacy risks. For example, the United States Department of Health and Human Services has made available a Synthetic Health Data Generation Engine\footnote{https://aspe.hhs.gov/synthetic-health-data-generation-engine-accelerate-patient-centered-outcomes-research} to accelerate patient-centred outcomes research and “address the need for research-quality synthetic data”. The United Kingdom’s National Health Service (NHS) has rolled out an `Artificial Data Pilot'\footnote{https://digital.nhs.uk/services/artificial-data} that aims to “provide users with large volumes of data that share some of the characteristics of real data while protecting patient confidentiality”. Similar efforts are recorded in Canada’s health economic hub Health City \cite{HealthCity} and Germany’s Charité Lab\footnote{https://claim.charite.de/en/} for Artificial Intelligence in Medicine.

Research-quality synthetic data (the focus of our work) can be used to rapidly develop and test preliminary hypotheses before applying them to real datasets \cite{articleKokosi}. They can also improve the research pipeline by acting as a proxy for real-world data \cite{articleGiuffre}. Furthermore, the controlled generation of synthetic health data can include a balanced representation of different demographic groups \cite{Micheletti2023.09.26.23296163}. This would ensure that the previously underrepresented socio-demographic groups are adequately represented, thereby mitigating biases in health research that arise from skewed real-world health datasets and, in turn, address model fairness \cite{unknownPreprint, articleBarbierato,unknownBreugel}.

\begin{wrapfigure}{l}{0.7\textwidth}
    \includegraphics[width=0.7\textwidth, trim={0cm 0cm 0cm 0.1cm},clip]{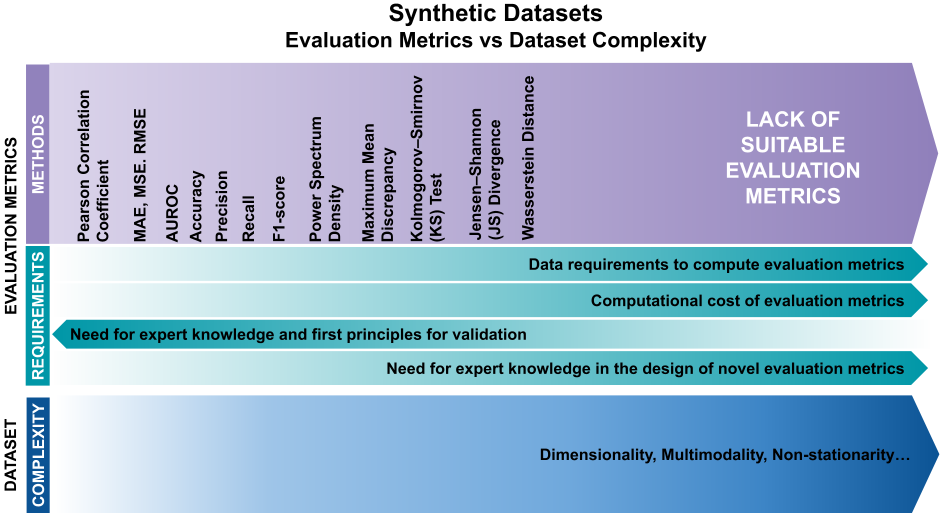} 
    \caption{There is an increasing lack of suitable evaluation metrics (due to increasing difficulties in computation and increasing difficulties in evaluation of the metrics), with the increasing complexity of synthetic datasets.}
    \label{fig:MainDiagram}
\end{wrapfigure}

However, despite the above-mentioned advantages of synthetic health data, major challenges remain with their large-scale adoption. One of the major challenge is the lack of consensus on evaluating synthetically generated data vis à vis the corresponding real data \cite{RodrguezAlmeida2022SyntheticPD,unknownJordon,unknownSchaar}. This not only makes it difficult to track the state-of-the-art progress of synthetic data generation methods but also poses barriers to trust and adoption, as well as presents regulatory and compliance challenges in safety-critical healthcare settings increasingly shaped by emerging governance expectations. 

To shed light on the evaluation approaches of synthetic data we have conducted a systematic review of $134$ research articles published in the last ten years. 

This is the first review of this type and size to understand which approaches are being used to evaluate the quality of synthetic data, along with the associated data generation methods and their target application areas. We provide a structured consolidation of generation and evaluation methods into taxonomies, alongside an analysis of their application contexts. We focused on structured tabular and time-series health data since this is one of the areas with the highest potential in advancing healthcare\cite{articleHernandez} and present unique data challenges, such as dealing with missingness. In addition, there is a higher consensus on the evaluation methods for other data modalities, such as imaging and text, where the respective research communities have developed metrics such as Fréchet Inception Distance (FID) \cite{inproceedingsFID} and BERTScore \cite{unknownBERTScore} respectively.

We observe that with the increasing complexity of the synthetic datasets (including dimensionality, multimodality, and non-stationarity), there is an increasing lack of suitable evaluation metrics, as shown in Fig.\ref{fig:MainDiagram}. This is manifested in the increasing difficulties in computation as well as evaluation of the metrics. Therefore, there's a critical need for the use of appropriate statistical evaluation metrics to critically evaluate complex synthetic data, including involvement of domain experts in: (i) selection of appropriate evaluation metrics and (ii) interpretation of the resulting outcomes. This type of collaboration between researchers and clinical practitioners can lead to the development of methods and metrics that implicitly incorporate domain knowledge, resulting in decreased need for expert knowledge in evaluating future synthetic data. As a result, distilling domain knowledge into operational constraints and ensuring that the underlying medical processes governing data generation are preserved will open the door to more robust and clinically meaningful machine learning evaluation paradigms.

In the following sections, we detail our methodology and the results of our analysis, followed by guidelines for evaluating synthetic tabular data, with the aim of supporting more standardised, transparent, and clinically meaningful evaluation practices.

\section{Methodology}

\begin{figure}
    \centering
    \includegraphics[width=0.87\textwidth, trim={0cm 0cm 0cm 0cm},clip]{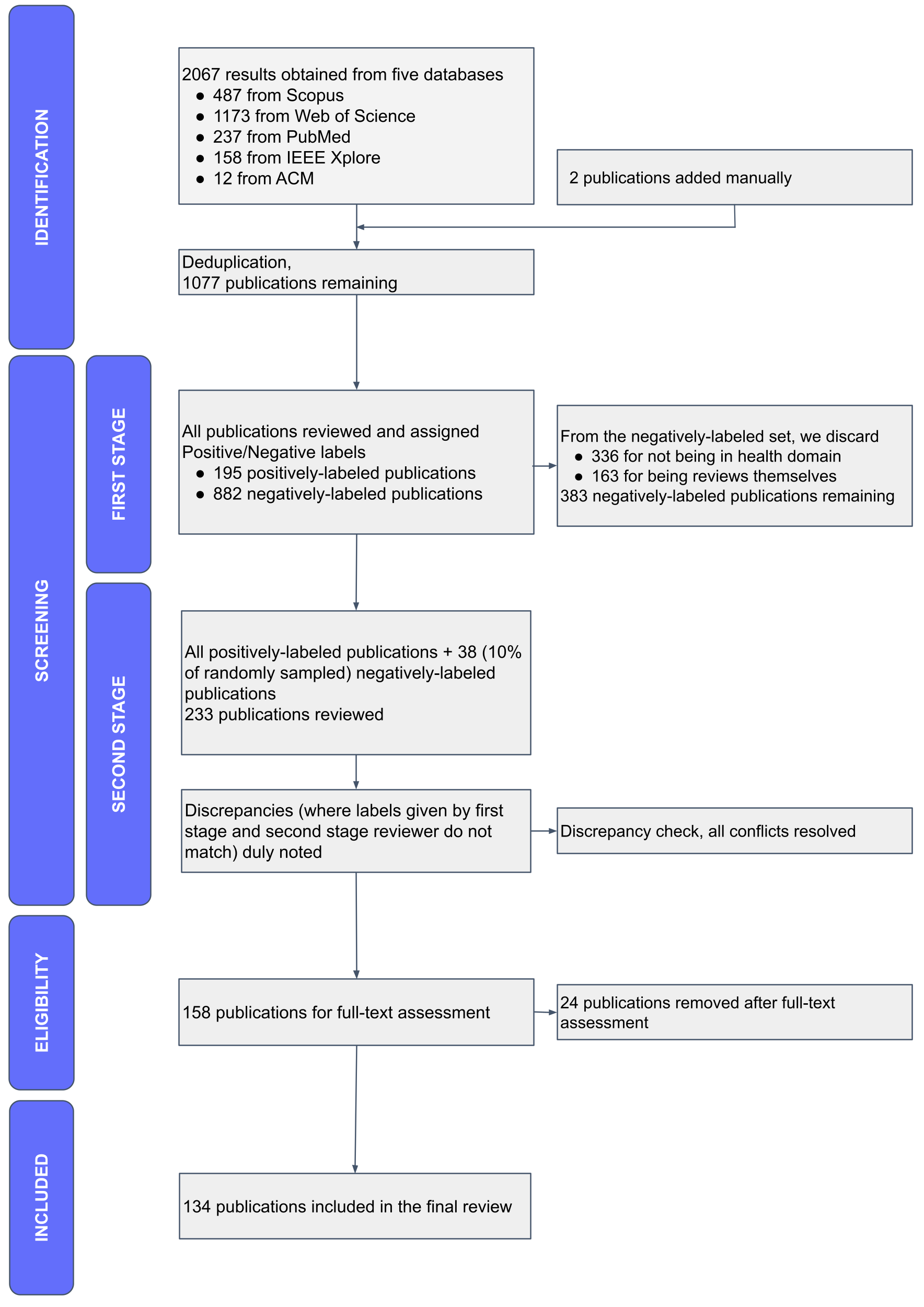}
    \caption{PRISMA flow diagram with details of the four phases: Identification, Screening, Eligibility, and Inclusion, as they relate to publications included in this review}
    \label{fig:prisma}
\end{figure}

In this section, we expand on our methodology for carrying out the systematic review. The Preferred Reporting Items for Systematic Reviews and Meta-Analyses (PRISMA) \cite{page2021prisma} statement provides a standardised framework for reporting systematic reviews. It consists of updated instructions on identifying, selecting, praising, and synthesising publications. Fig.\ref{fig:prisma} outlines our methodology in accordance with the latest PRISMA guidelines. 

\subsection{Search Strategy} \label{SearchStrategy}

To establish an unambiguous search strategy, we laid out the following: \textbf{(i.)} The relevant databases used to search within a time frame, \textbf{(ii.)} The search terms to ensure comprehensive coverage of relevant studies, and, \textbf{(iii.)} The inclusion and exclusion criteria. Additional details about the search strategy, including Search Queries, Selection Process, and Data Items, are reported in the Appendix (\ref{appendix:searchqueries}).\\

\textbf{Databases and Search Engines:} For this systematic review, we looked for publications on the following five databases: Scopus\footnote{https://www.scopus.com}, Web of Science\footnote{https://www.webofscience.com}, PubMed\footnote{https://pubmed.ncbi.nlm.nih.gov}, IEEE Xplore\footnote{https://ieeexplore.ieee.org}, and Association for Computing Machinery (ACM)\footnote{https://www.acm.org}. Additional publications were also manually selected using Google Scholar.\\

\textbf{Search Terms and Additional Limits:} \textit{Search Terms}: We began by identifying publications dealing with synthetic data generation or augmentation. The focus was on finding publications that dealt with tabular or time-series data using wildcards in the search, and included all similar words, including plurals and noun and verb forms of words. Keywords such as 'patient', 'health', and 'clinical' were also used to conduct the search within the health domain. As a result of the aforementioned considerations, the following search string was designed: 
\textit{(synthe* OR augment*) AND generat* AND (time-series OR time* OR temporal*) AND (tabular OR record*) AND (patient* OR medic* OR health* OR clinic* OR ehr*)}, to be searched in the title-abstract-keywords or the topic field. \textit{Additional Limits}: We limited our search to publications in the last ten years. We also limited the search to peer-reviewed conferences and journal articles written in the English language.\\

\textbf{Inclusion and Exclusion Criteria:} We used a set of criteria for inclusion or exclusion of publications in this review. \textit{Inclusion Criteria}: The publications that were included in this systematic review met the following conditions:

\begin{itemize}
\item Publications that deal with tabular or time-series data.
\item Publications that describe a method of generation of synthetic data and its evaluation against real data, or Publications that do not describe a method of generation of synthetic data but describe its evaluation vis-à-vis real data.
\item Publications that deal with the generation of completely new synthetic datasets, as well as those that deal with the augmentation of existing datasets with synthetic data.
\item Peer-reviewed publications from journals and conferences. Strictly no pre-prints.
\end{itemize}

\textit{Exclusion Criteria}: All possible publications that would be irrelevant to our study were excluded if they met any one of the following conditions:

\begin{itemize}
\item Publications that are not in the health domain.
\item Publications that deal with image-, audio-, video-, or text-only modalities of data.
\item Publications which themselves are narrative or systematic reviews.
\item Publications on synthetic data that neither describe a method of generation of synthetic data nor its evaluation against real data.
\item Publications that describe a method of generation of synthetic data, but the synthetic data is not structurally similar (and therefore, not comparable) to real data.
\end{itemize}

\subsection{Reporting}

The reporting of this systematic review adheres to the PRISMA\cite{page2021prisma} guidelines. We undertook measures to ensure transparency and reproducibility and facilitate critical appraisal and interpretation of the findings. \\

\textbf{Risk of Bias Management:} To establish the transparency of the findings and the results of this systematic review, we: (i.) used multiple databases to ensure no platform-specific bias creeps in, (ii.) used value-neutral search terms in the search query (iii.) got the publications reviewed by five reviewers in multiple screening stages (iv.) performed spot checks and discrepancy checks to ensure no reviewer-induced bias creeps in.

\section{Results}
Based on a detailed review of $134$ publications, we present the following results, grouped into four categories: Evaluation, Generation, Purpose and Impact of synthetic data including the reproducibility of the results.

\subsection{Evaluation of Synthetic Data} 

We categorise the approaches used in the evaluation of synthetic data into: (i) Direct vs Indirect approaches, (ii) Quantitative vs Qualitative methods, and (iii) Utility, Fidelity, and Privacy taxonomy.

\subsection*{(i) Direct vs Indirect approaches} 

\begin{figure}[htbp]
    \centering
    \begin{subfigure}[t]{0.45\textwidth}
        \centering
        \includegraphics[width=0.83\textwidth]{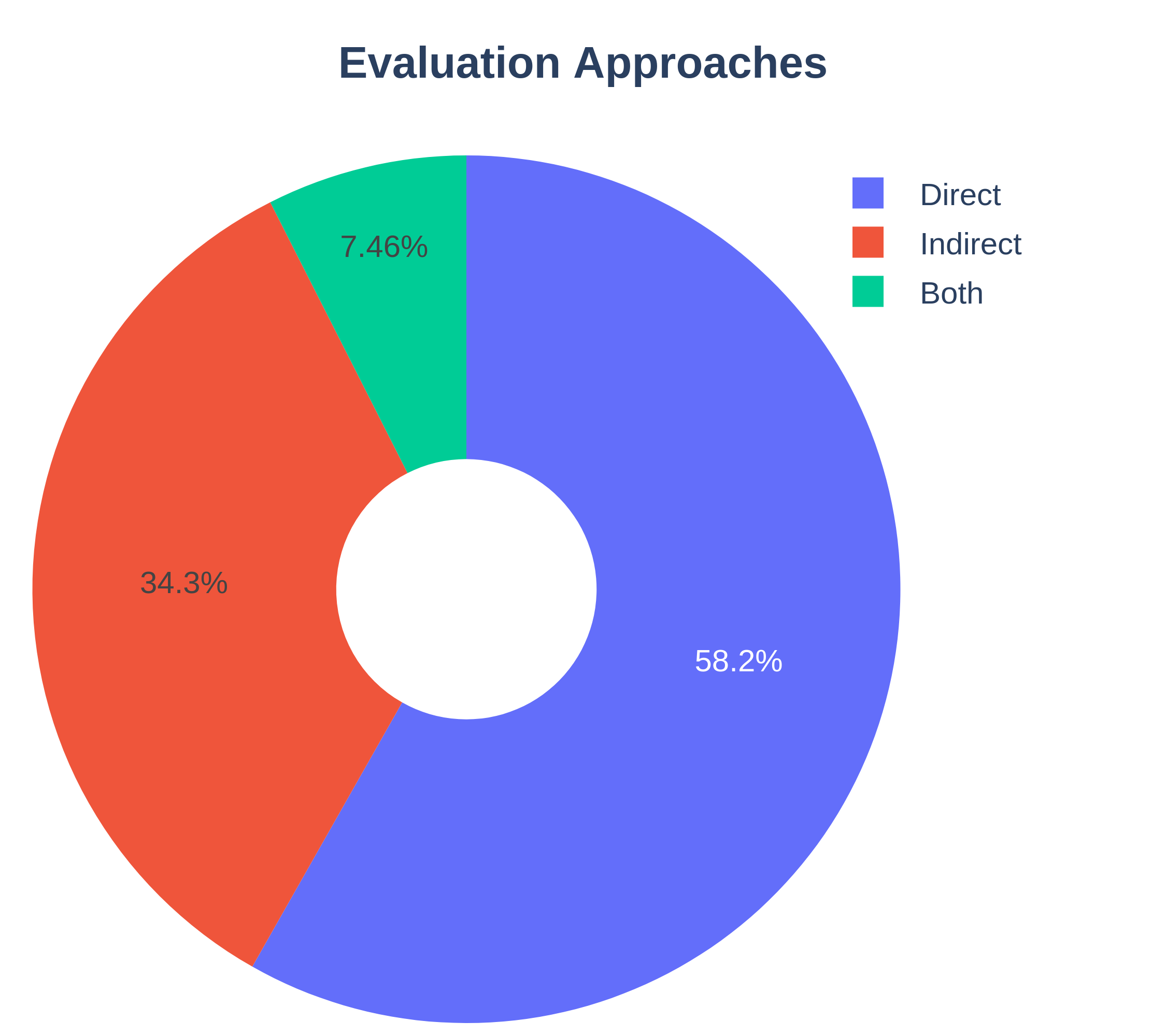}
        \caption{Breakdown of Evaluation Approaches into Direct vs Indirect Evaluation}
        \label{fig:Evaluation_Methods_2026_a}
    \end{subfigure}
    \hfill
    \begin{subfigure}[t]{0.45\textwidth}
        \centering
        \includegraphics[width=0.9\textwidth]{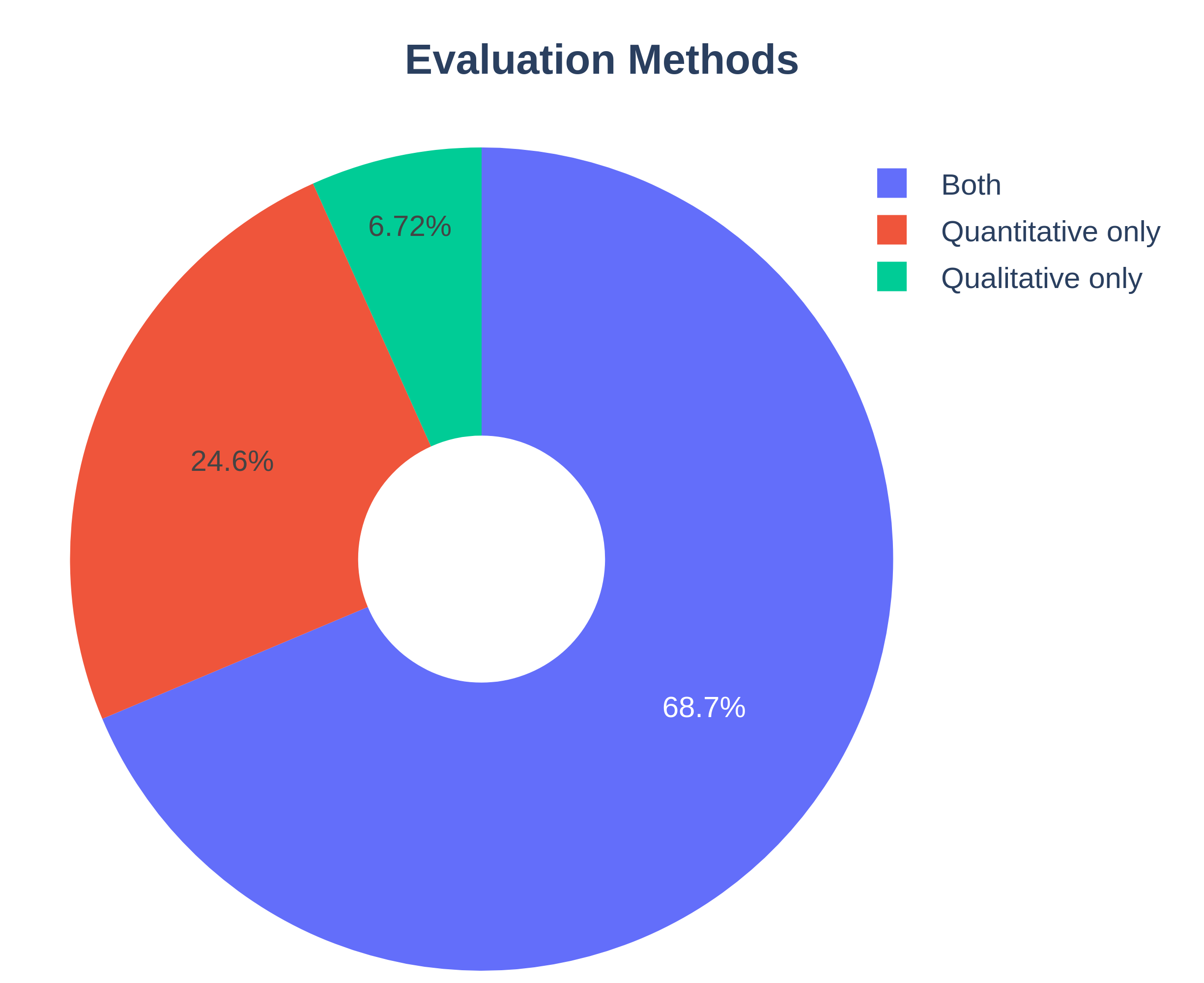}
        \caption{Breakdown of Evaluation Methods into
Quantitative vs Qualitative Methods}
        \label{fig:Evaluation_Methods_2026_b}
    \end{subfigure}
    \caption{Breakdown of Evaluation Approaches and Evaluation Methods}
    \label{fig:Evaluation_Methods_2026}
\end{figure}

\textbf{Direct evaluation approaches} involve using existing, standardised metrics to assess the quality of the synthetic data. \textbf{Indirect evaluation approaches} include non-standardised, domain-specific methods (such as TSTR - Train on Synthetic, Test on Real) to assess synthetic data in real-world applications. Indirect approaches extend beyond the standard metrics and often include context-driven or subjective evaluations. We found that Direct approaches are the most common at $58.2$\%. About $34.3$\% of the publications use Indirect approaches. Additionally, $7.46$\% use both Direct and Indirect approaches in conjunction, to perform a holistic evaluation of their synthetic data (Fig. \ref{fig:Evaluation_Methods_2026_a}).

\subsection*{(ii.) Quantitative vs Qualitative methods}

\begin{figure}[htbp]
    \centering
    \begin{subfigure}[t]{0.48\textwidth}
        \centering
        \includegraphics[width=0.85\textwidth]{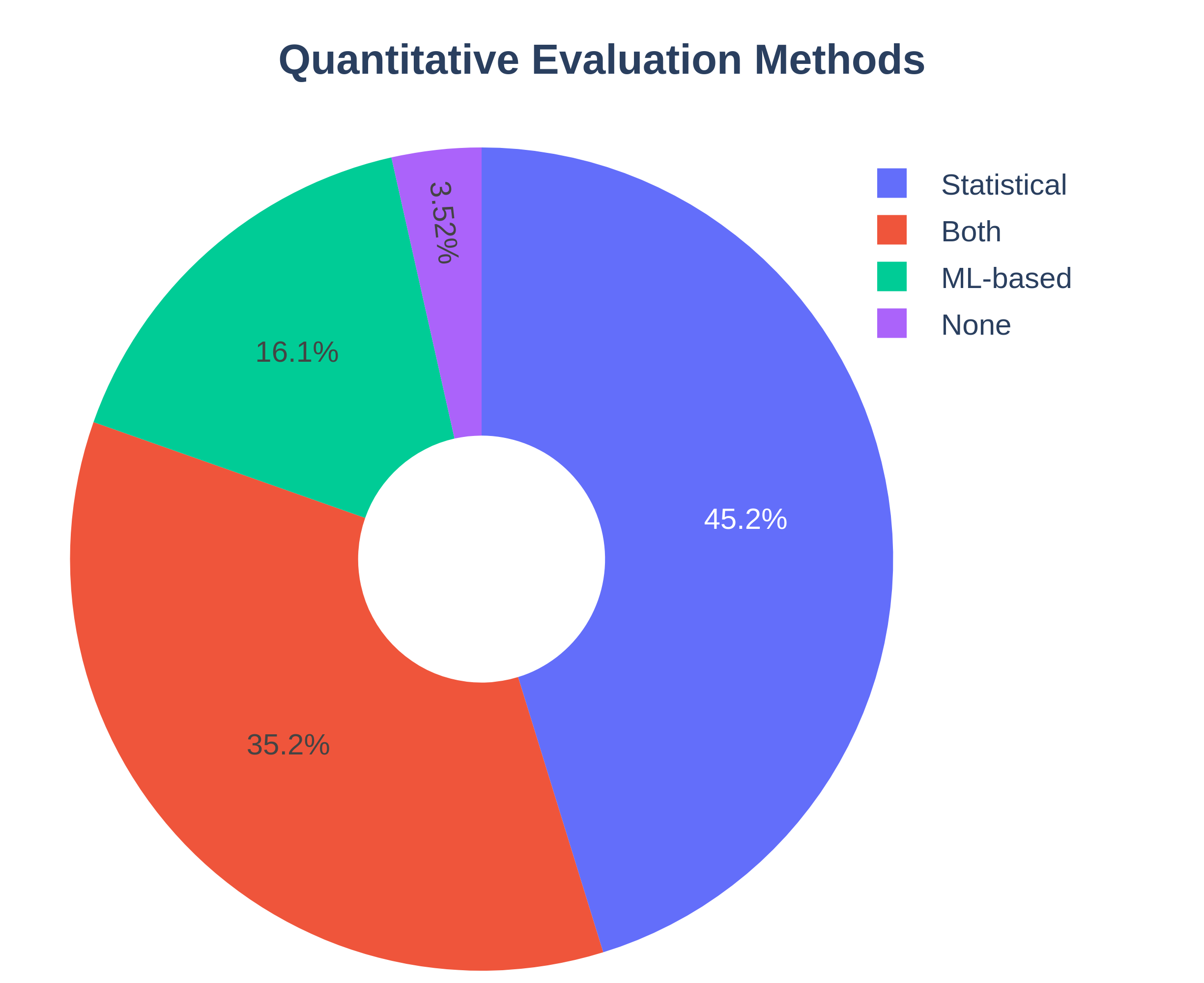}
        \caption{Breakdown of Quantitative Evaluation Methods into Statistical and ML-based methods}
        \label{fig:Qualitative_Evaluation_Methods_2026_a}
    \end{subfigure}
    \hfill
    \begin{subfigure}[t]{0.48\textwidth}
        \centering
        \includegraphics[width=0.85\textwidth]{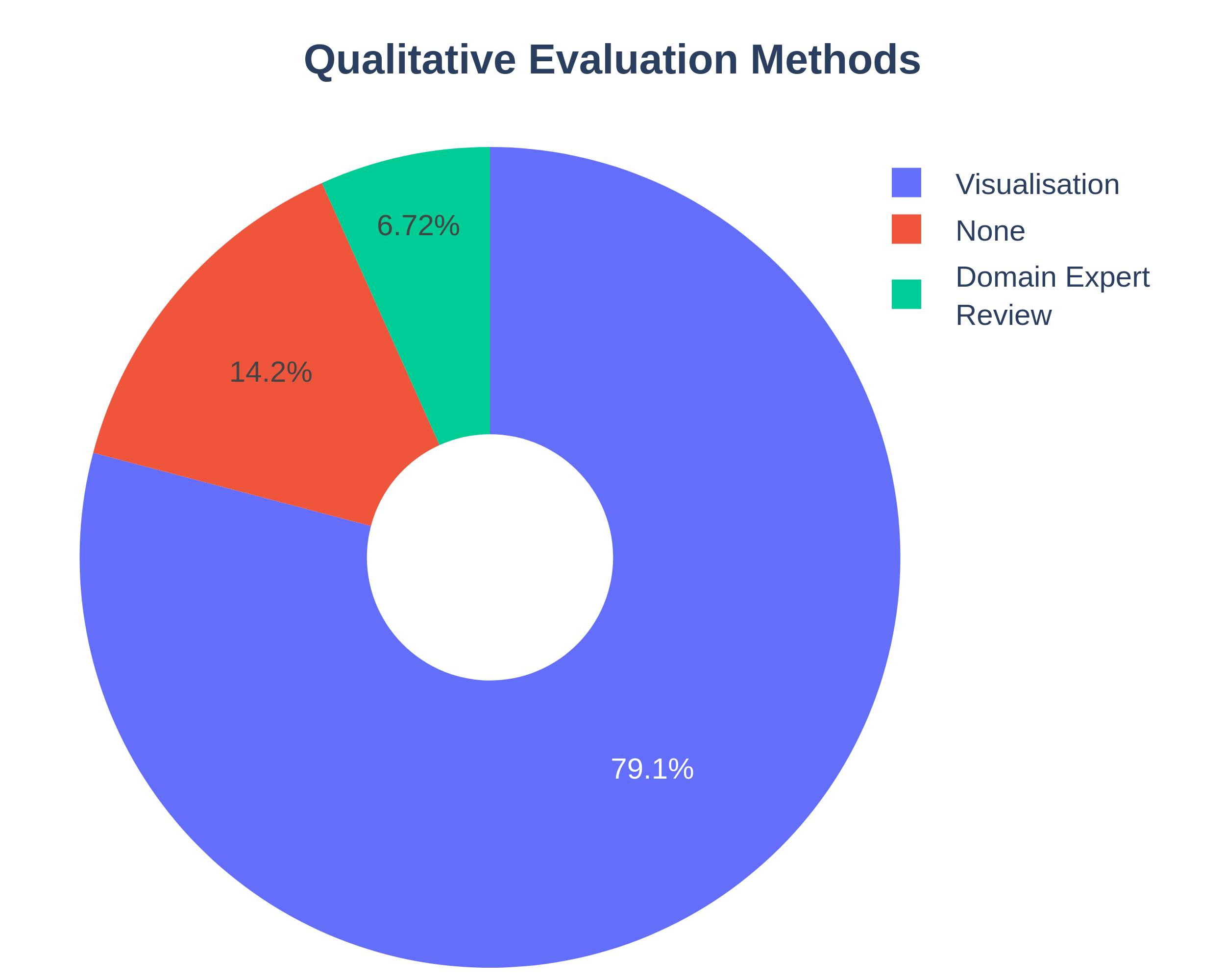}
        \caption{Breakdown of Qualitative Evaluation Methods into Visualisation and Domain Expert Review}
        \label{fig:Qualitative_Evaluation_Methods_2026_b}
    \end{subfigure}
    \caption{Breakdown of Quantitative and Qualitative Evaluation Methods}
    \label{fig:Qualitative_Evaluation_Methods_2026}
\end{figure}

\textbf{Quantitative evaluation methods} give objective, measurable results and are crucial for ensuring that synthetic data aligns statistically with real data. They include \textbf{Statistical techniques}, which use quantifiable metrics to compare synthetic data with the original data, and \textbf{ML-based techniques}, which make use of classification and/or regression to assess how well synthetic data performs when used for specific downstream tasks. We found that Statistical evaluation is the most popular ($45.2$\%), whereas ML-based techniques feature in $16.1\%$ of all publications. About $35.2\%$ publications use both Statistical and ML-based evaluation techniques in conjunction. (Fig. \ref{fig:Qualitative_Evaluation_Methods_2026_a}).

\begin{figure}[htbp]
    \centering
    \includegraphics[width=0.6\textwidth, trim={0cm 0cm 0cm 0cm},clip]{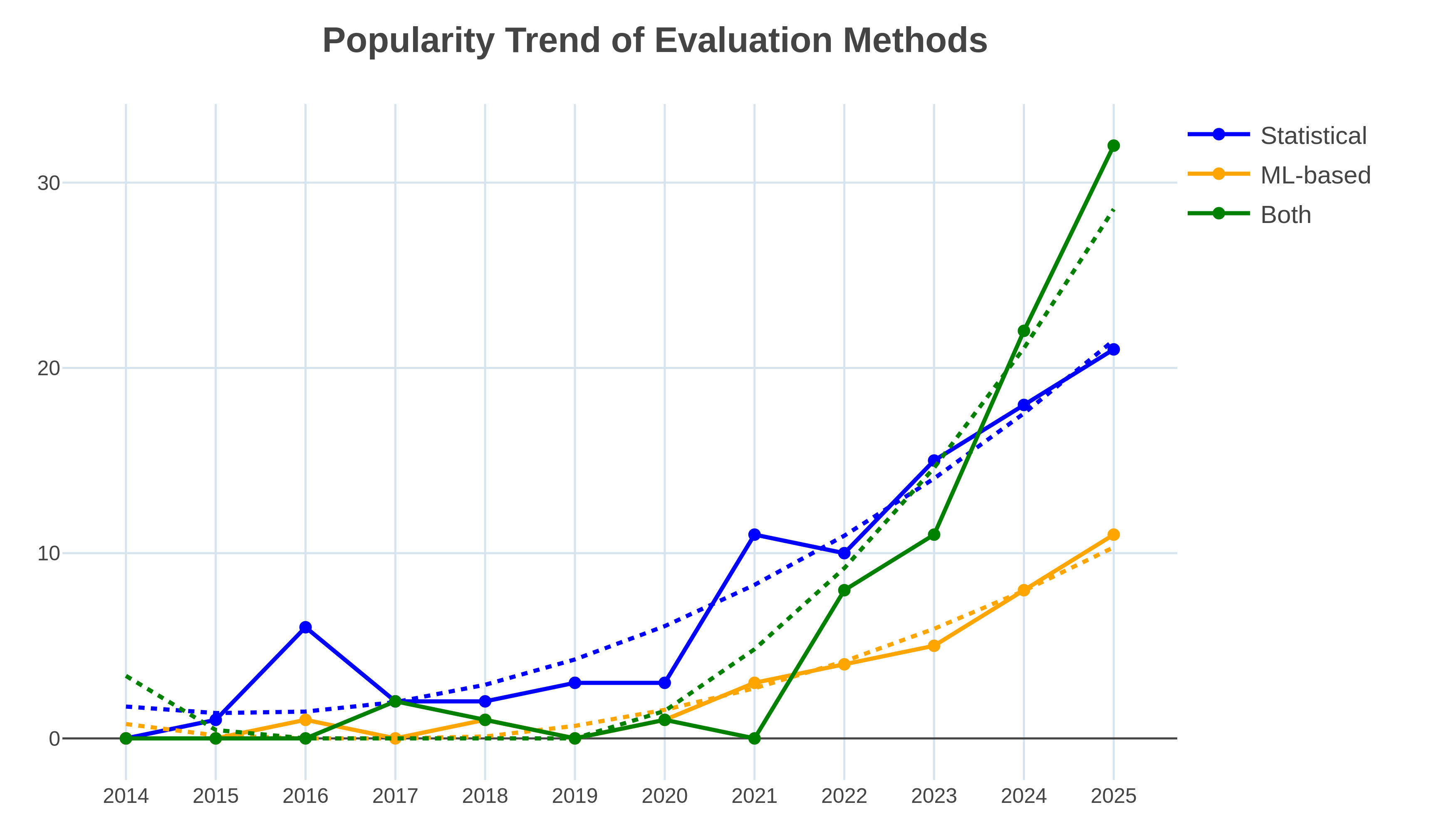}
    \caption{Popularity trend of Statistical and ML-based Evaluation Methods over the last decade, as obtained from publications included in this review.  X-axis represents the year, Y-axis represents the number of publications. Dotted lines represent the overall polynomial trend.}
    \label{fig:Popularity_Trend_of_Evaluation_Methods_2026}
\end{figure}

The popularity of Statistical evaluation techniques remains high; however, a trend can be seen in publications using both Statistical and ML-based techniques together (Fig. \ref{fig:Popularity_Trend_of_Evaluation_Methods_2026}). 

\begin{figure}[htbp]
    \centering
    \begin{subfigure}[t]{0.5\textwidth}
        \centering
        \includegraphics[width=1\textwidth]{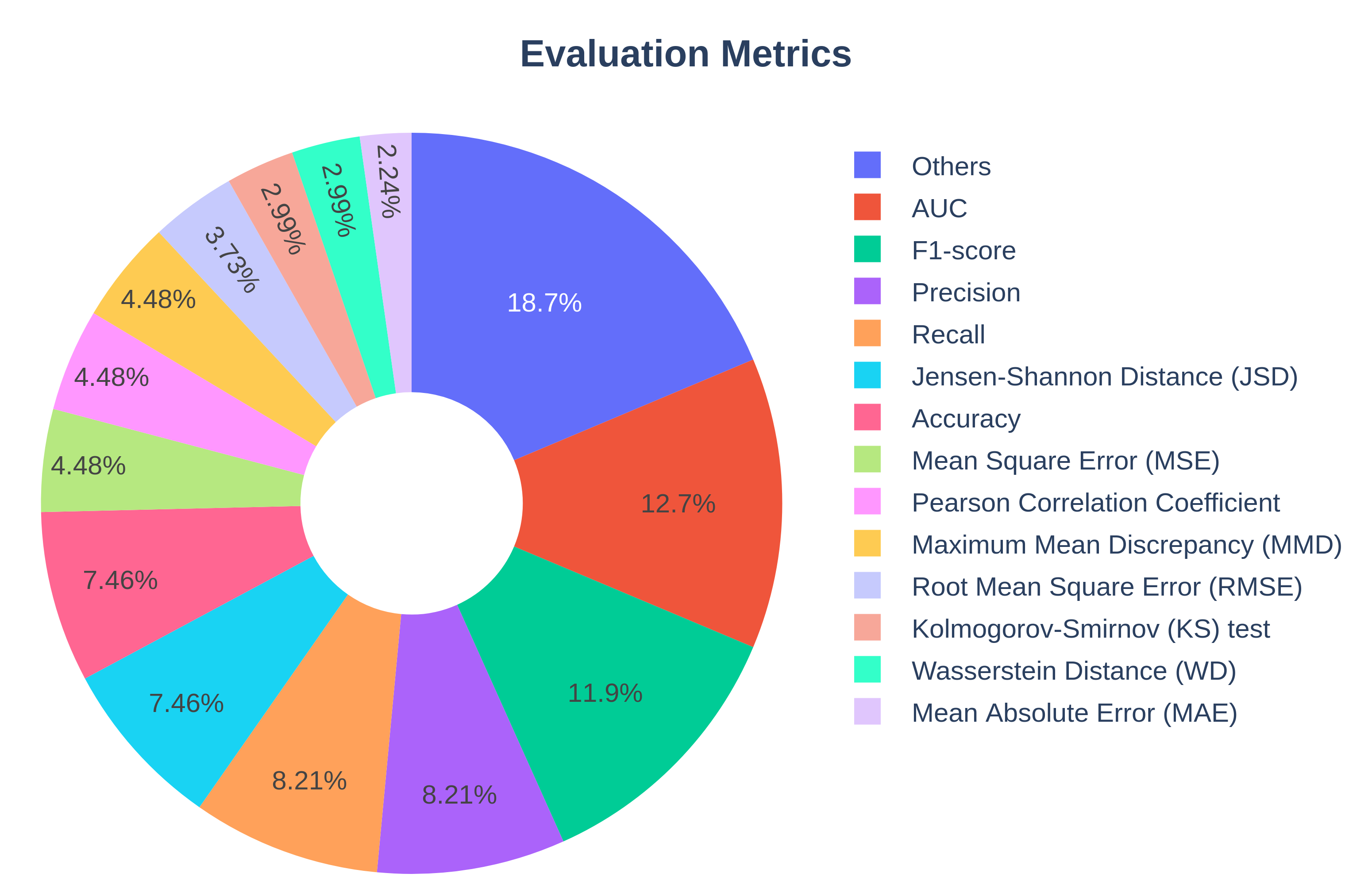}
        \caption{Most popular Evaluation Metrics}
        \label{fig:Author_defined_Metrics_2026_a}
    \end{subfigure}
    \hfill
    \begin{subfigure}[t]{0.4\textwidth}
        \centering
        \includegraphics[width=0.9\textwidth]{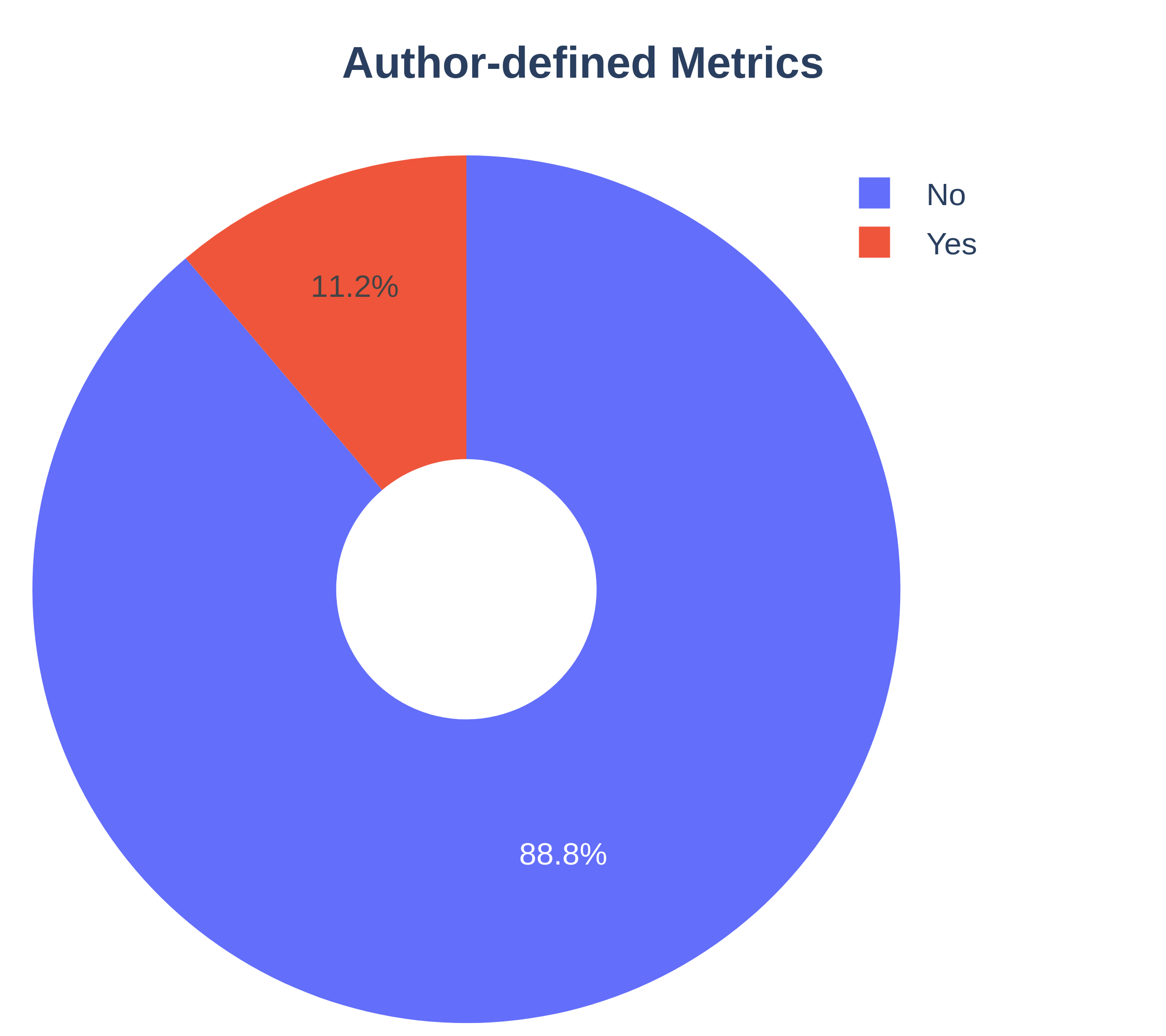}
        \caption{Breakdown of whether the metric is author-defined or not}
        \label{fig:Author_defined_Metrics_2026_b}
    \end{subfigure}
    \caption{Breakdown of Evaluation Metrics}
    \label{fig:Author_defined_Metrics_2026}
\end{figure}

The most used quantitative evaluation metrics include Jensen-Shannon (JS) distance, Pearson Correlation coefficient, and Maximum Mean Discrepancy (MMD), apart from the popular metrics such as AUC and F1-score for classification tasks, and Mean Squared Error (MSE) and Root Mean Squared Error (RMSE) for regression tasks (Fig. \ref{fig:Author_defined_Metrics_2026_a}). We also note that the majority of the included papers (88.8\%) use existing metrics, and only 11.2\% of the publications use their own Author-defined metric for evaluation of the synthetic data (Fig. \ref{fig:Author_defined_Metrics_2026_b}).

\textbf{Qualitative evaluation methods} rely on subjective judgement and human interpretation to assess the quality of synthetic data. In the majority of the cases (79.1\%), they take the form of \textbf{Visualisation} of the graphical representation of distributions of synthetic data (Fig. \ref{fig:Qualitative_Evaluation_Methods_2026_b}). We also note that despite its significance, use of \textbf{Domain Expert Review} as a qualitative evaluation method is not yet widely used, and was found only in a handful of papers (6.72\%).

\begin{figure}
    \includegraphics[width=0.85\textwidth, trim={0cm 2cm 3cm 0cm},left,clip]{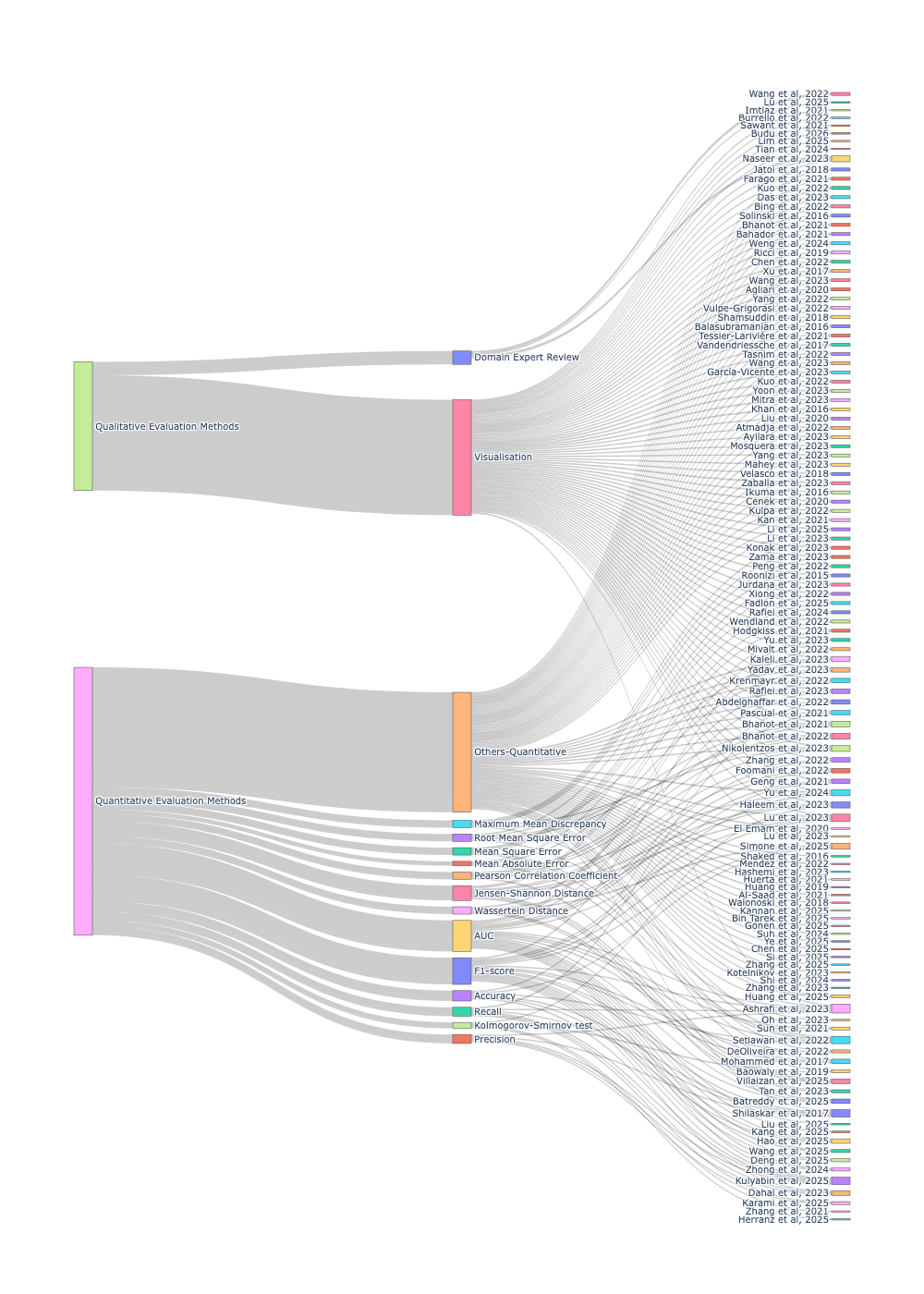}
    \caption{Sankey diagram depicting the most popular Evaluation Metrics and the papers utilising them, as obtained from the publications included in this review}
    \label{fig:Sankey}
\end{figure}

Quantitative methods by themselves are used in about $24.6$\% of all publications, whereas for qualitative, this value is $6.72$\%. Fig. \ref{fig:Evaluation_Methods_2026_b} shows that most research ($68.7$\%) uses a combination of both quantitative and qualitative methods. Fig. \ref{fig:Sankey} gives a detailed depiction of the most popular evaluation metrics and the papers utilising them that have been included in this review. 

\subsection*{(iii.) Utility, Fidelity, Privacy Taxonomy}

The Utility, Fidelity, and Privacy framework categorises evaluation methods under the following labels: \textbf{Utility} refers to the ability of synthetic data to support downstream analytical or predictive tasks, \textbf{Fidelity} captures the statistical resemblance of synthetic data to real data, and \textbf{Privacy} reflects the risk of sensitive information leakage. The resulting taxonomy, summarised in Table~\ref{tab:evaluation_methods_taxonomy}, provides a concise and unified view of evaluation methods across the literature according to this framework. 

It should be noted, however, that there isn’t always a strict one-to-one mapping between individual evaluation methods and these three umbrella terms; some methods/metrics may capture multiple aspects simultaneously or be interpreted differently depending on context. Therefore, this categorisation should be viewed as a guiding framework rather than an exhaustive or rigid taxonomy, serving to navigate the landscape of evaluation rather than to strictly confine each method.

\begin{table*}[t]
\centering
\footnotesize
\begin{tabularx}{\textwidth}{p{1.5cm} p{2.25cm} X}
\toprule
\textbf{Category} & \textbf{Subcategory} & \textbf{Methods and Metrics} \\
\midrule
\multirow{4}{*}{\textbf{Utility}} 
& Evaluation Strategies 
& Train on Synthetic, Test on Real (TSTR); Train on Real, Test on Synthetic (TRTS) \\

& Predictive Models 
& Logistic Regression (LR), Random Forest (RF), Support Vector Machine (SVM), K-Nearest Neighbours (KNN), Decision Trees, Naive Bayes, XGBoost, Neural Networks (MLP, CNN, RNN, LSTM) \\

& Performance Metrics 
& Accuracy, Precision, Recall (Sensitivity), Specificity, F1-score, AUROC (AUC), AUPRC, Cohen’s Kappa, Error Rate \\

& Human Evaluation 
& Expert/clinician review, Participant satisfaction questionnaires \\

\midrule
\multirow{5}{*}{\textbf{Fidelity}} 
& Similarity 
& Jensen--Shannon Divergence (JSD), Kullback--Leibler Divergence (KL), Wasserstein Distance, Hellinger Distance, Bhattacharyya Distance, Maximum Mean Discrepancy (MMD), Graph Kernel MMD (GK-MMD), Weisfeiler--Lehman kernel, Shortest-path kernel, PCA, t-SNE, UMAP, Nearest neighbour similarity, Pairwise similarity scores, Sample matching, Log-likelihood, ECG morphology similarity, channel/spatial covariance, disease distribution similarity \\

& Statistical Tests 
& Kolmogorov--Smirnov (KS) test, Mann--Whitney U test, Wilcoxon rank-sum test, Benjamini--Hochberg procedure \\

& Error Metrics 
& Mean Squared Error (MSE), Root Mean Squared Error (RMSE), Mean Absolute Error (MAE), Per cent Root Difference (PRD) \\

& Distance Metrics 
& Euclidean Distance, Dynamic Time Warping (DTW), Fréchet Distance (FD) \\

& Correlation 
& Pearson Correlation, Pairwise Correlation, Autocorrelation Function (ACF), Concordance Correlation Coefficient (CCC) \\

\midrule
\multirow{2}{*}{\textbf{Privacy}} 
& Disclosure 
& Membership inference attacks, Presence disclosure risk, Attribute inference attacks \\

& Memorisation Risk 
& Nearest neighbour adversarial accuracy, k-NN memorisation analysis, Attacker Advantage (AA), Differential Privacy \\
\bottomrule
\end{tabularx}
\caption{Taxonomy of Evaluation Methods: Utility, Fidelity, and Privacy}
\label{tab:evaluation_methods_taxonomy}
\end{table*}

\subsection{Generation of Synthetic Data}

We categorise the models used for the generation of synthetic data into Probabilistic vs Mechanistic models.

\begin{figure}[h]
    \centering
    \begin{subfigure}[t]{0.48\textwidth}
        \centering
        \includegraphics[width=0.8\textwidth]{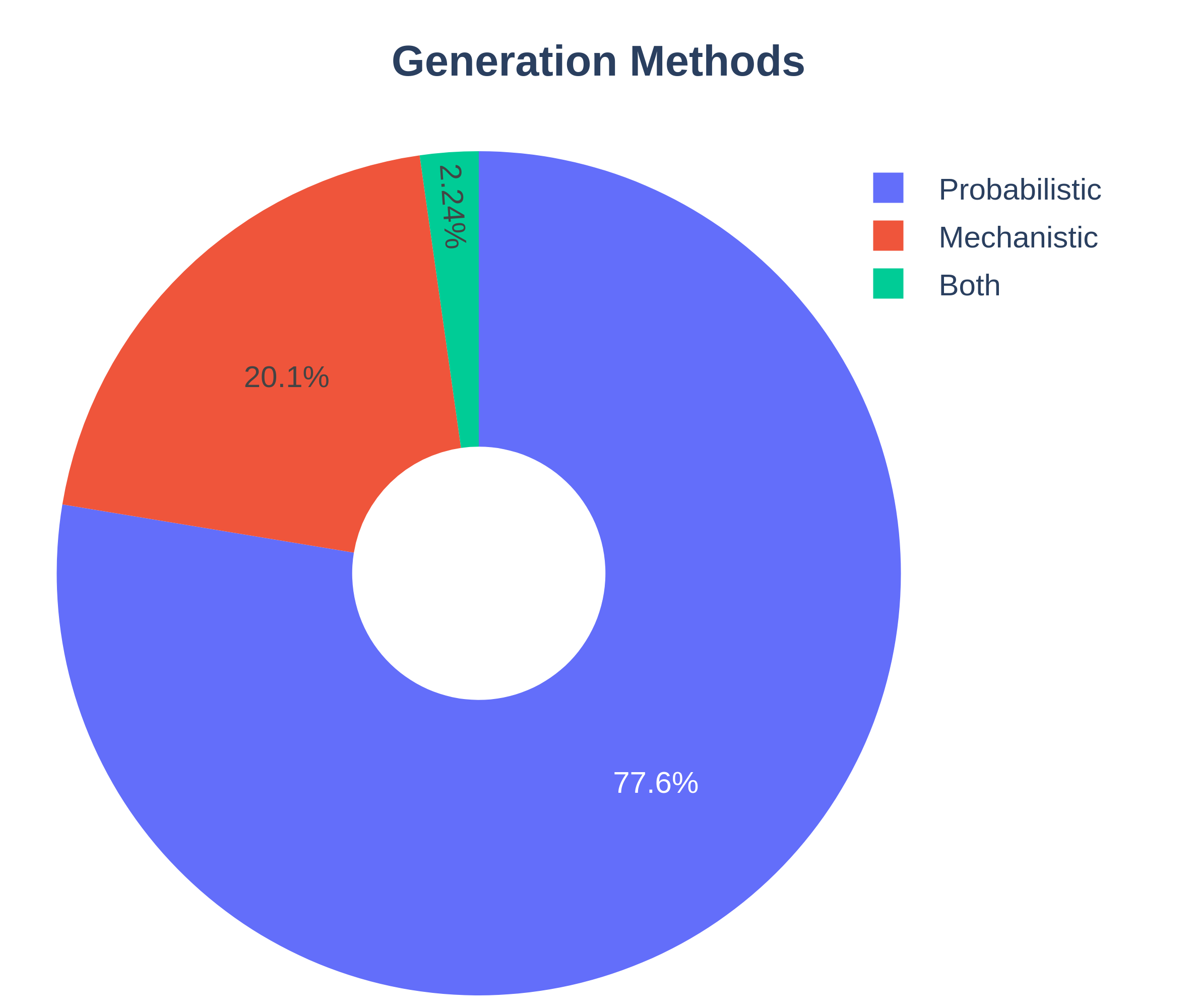}
        \caption{Breakdown of Generation Methods into Probabilistic vs Mechanistic Models}
        \label{fig:Generation_Models_2026_a}
    \end{subfigure}
    \hfill
    \begin{subfigure}[t]{0.48\textwidth}
        \centering
        \includegraphics[width=1\textwidth]{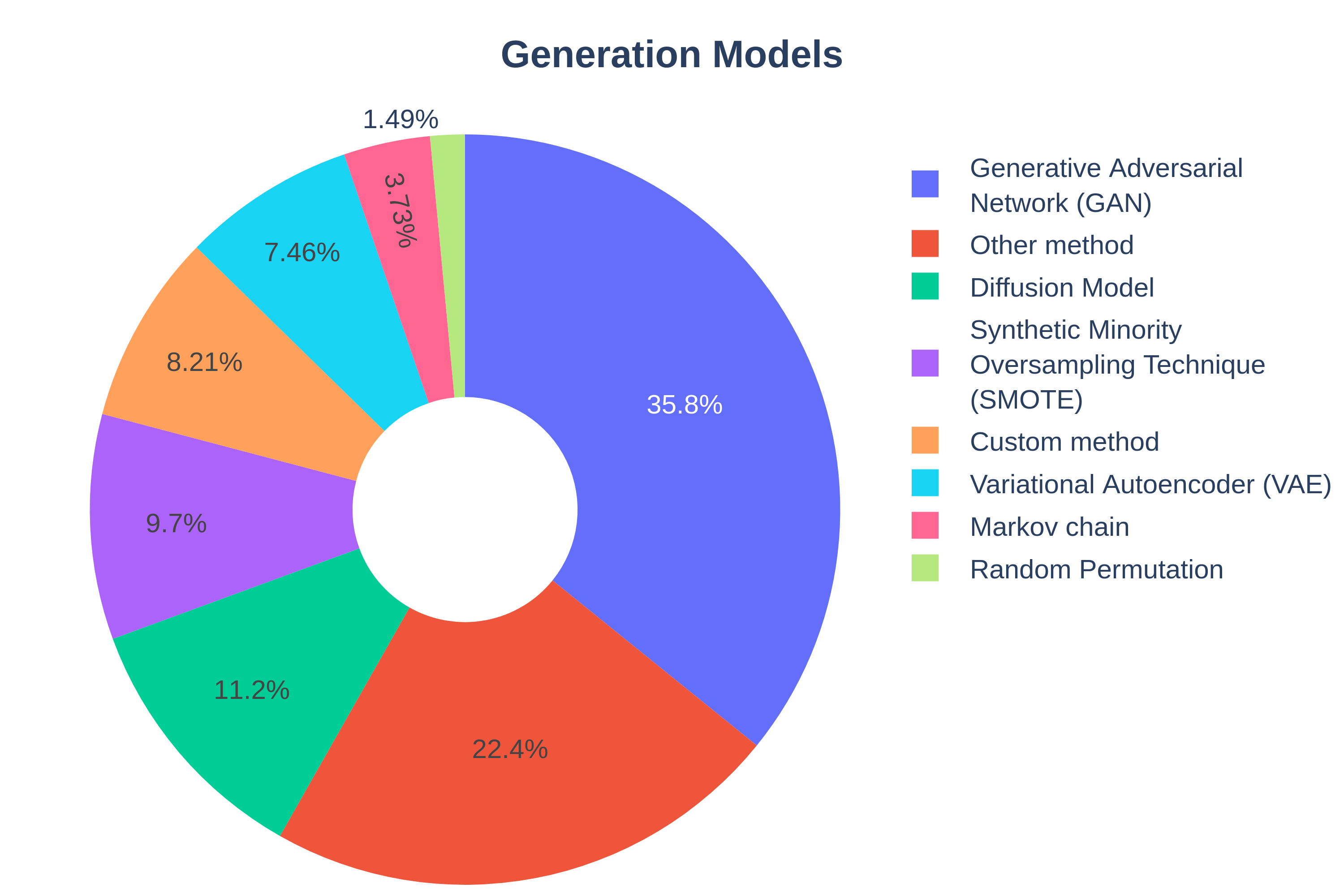}
        \caption{Most popular Generation Models}
        \label{fig:Generation_Models_2026_b}
    \end{subfigure}
    \caption{Breakdown of Generation Models}
    \label{fig:Generation_Models_2026}
\end{figure}

\subsection*{(i.) Probabilistic vs Mechanistic models}

\textbf{Probabilistic Models} use statistical and probability distribution approaches to capture the statistical properties (such as distribution, correlations, and relationships between variables) of the real data, to generate the synthetic data. \textbf{Mechanistic Models}, on the other hand, use explicit rules, equations, or processes to simulate data based on how the underlying systems work. The most popular generation models are based on GANs, SMOTE, VAEs, Markov Chains, and Random Permutations (Fig. \ref{fig:Author_defined_Metrics_2026_a}). Diffusion-based models are seeing a rise in popularity for longitudinal tabular data.

\begin{figure}[h]
    \centering
    \includegraphics[width=0.6\textwidth, trim={0cm 0cm 0cm 0cm},clip]{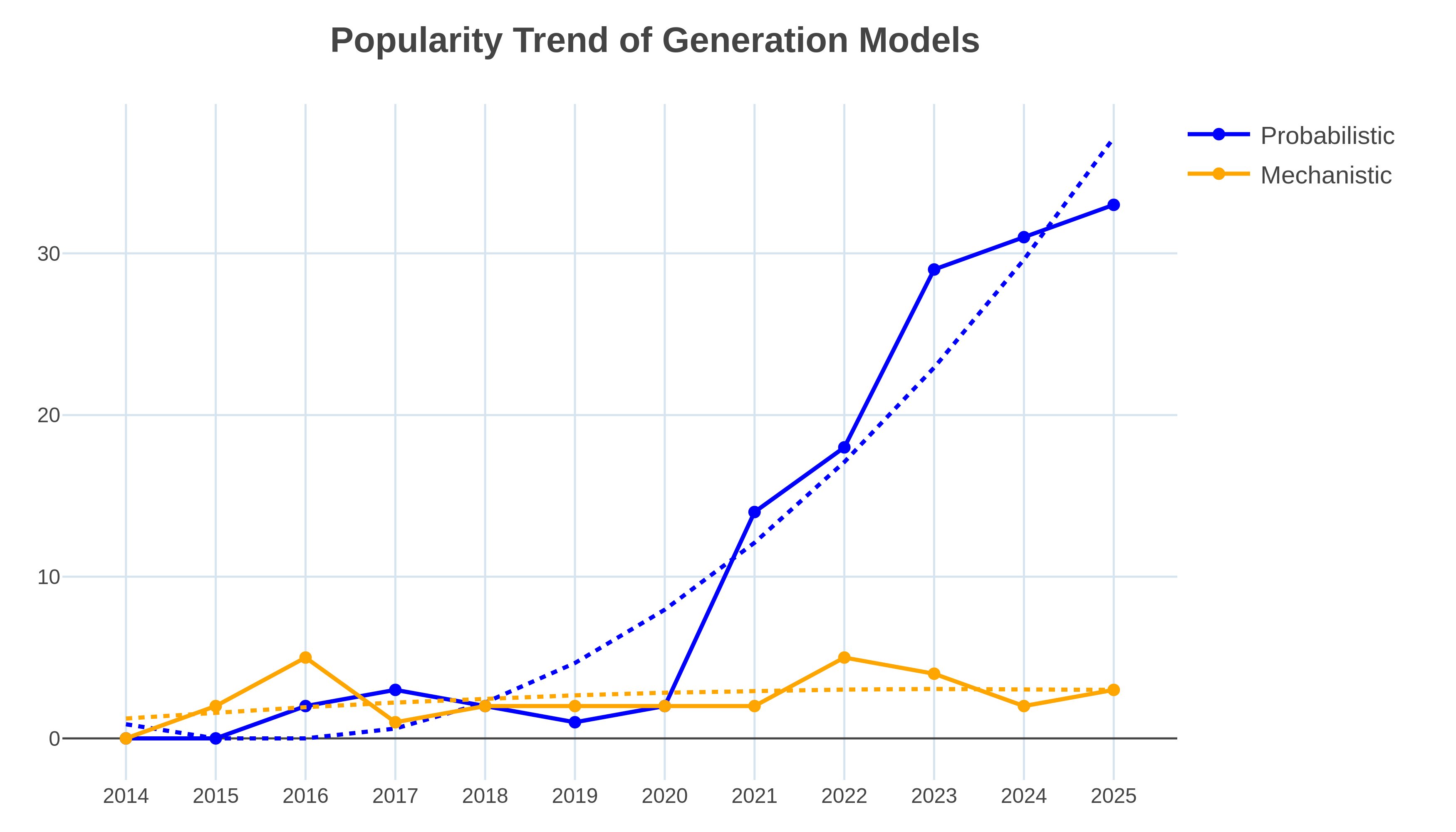}
    \caption{Popularity trend of Probabilistic and Mechanistic Generation Models over the last decade, as obtained from publications included in this review. X-axis represents the year, Y-axis represents the number of publications.  Dotted lines represent the overall polynomial trend.}
    \label{fig:Popularity_Trend_of_Generation_Models_2026}
\end{figure}

We also observed a growing divergence between Probabilistic and Mechanistic models, with Probabilistic Models increasingly being more frequently used (77.6\%) and Mechanistic Models tending to be more referenced in older publications (Fig. \ref{fig:Popularity_Trend_of_Generation_Models_2026}).

\subsection{Purpose and Impact of Synthetic Data} 

\begin{wrapfigure}{l}{0.6\textwidth}
    \includegraphics[width=0.6\textwidth, trim={0cm 0cm 0cm 0cm},clip]{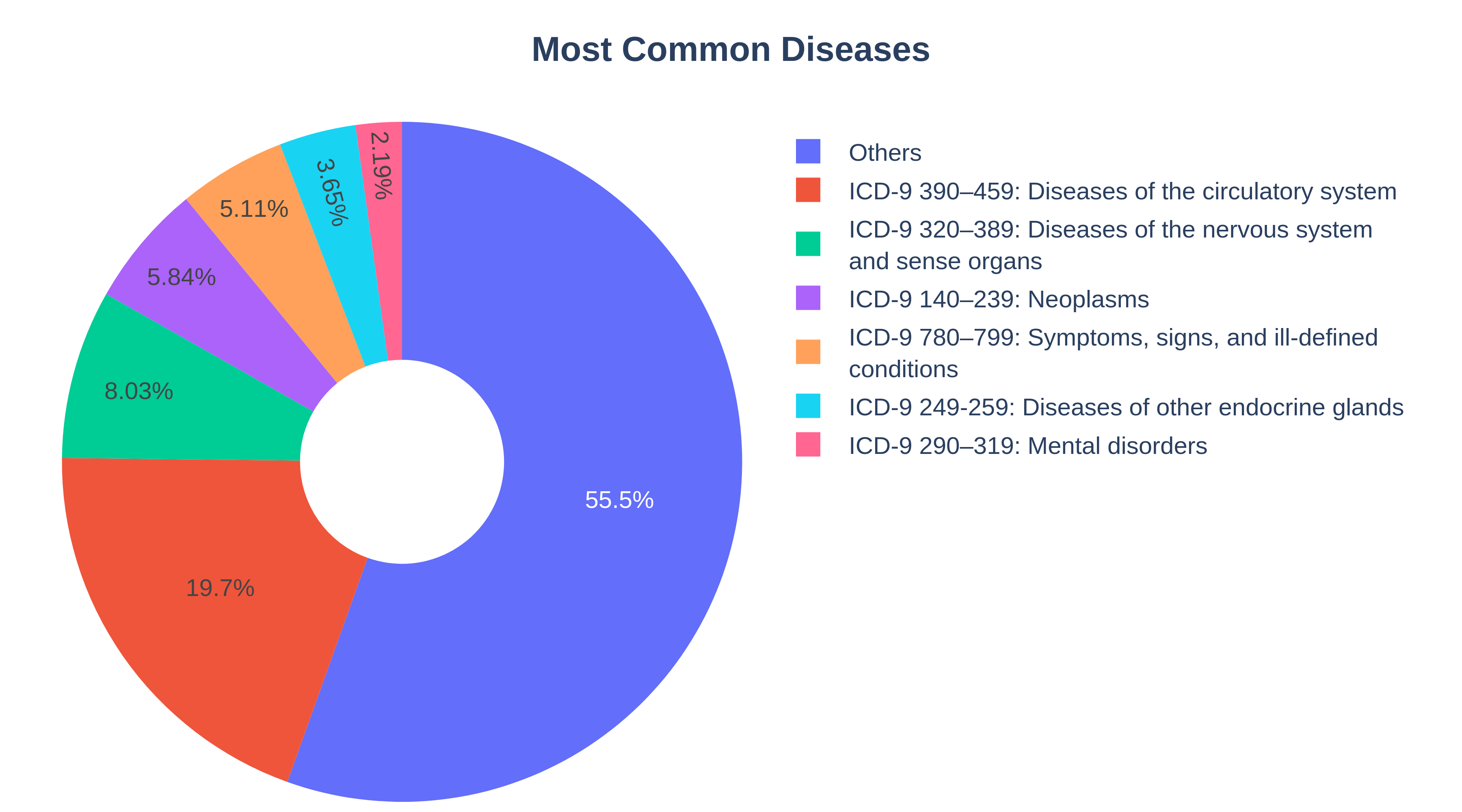}
    \caption{Most common diseases (grouped by their ICD-9 codes), as seen through the publications included in this review.
    }
    \label{fig:Most_Common_Diseases_2026}
\end{wrapfigure}

We found that privacy preservation, followed by predictive modelling and data-quality enhancement, are the most popular objectives for the use of synthetic tabular data in healthcare. The most common diseases within the set of publications included in this review, for which synthetic health data are used, include those of the circulatory system, the nervous system, endocrine glands, and neoplasms (Fig. \ref{fig:Most_Common_Diseases_2026}). This may be driven by the popularity of the datasets, with MIMIC III\cite{johnson2016mimic} and MIMIC IV\cite{johnson2023mimic} being the most popular tabular and time-series health datasets.

The most widely-used repositories are Physionet\footnote{https://physionet.org/} and UCI ML Repository\footnote{https://archive.ics.uci.edu/datasets/}, where most datasets are held, a testament to the value they provide to the research community. At the same time, this also poses a risk of perpetuating possible biases that may be present in the resources themselves (for example, underrepresentation bias) to a global user base.

While there is some evidence to suggest that clinically oriented journals are also beginning to consider synthetic health data research, the majority of the articles are published in journals and conferences with a primarily technical focus. Most of the research in synthetic health data is carried out in North America, followed by Europe and Asia, which may be influenced by the availability of health data and data protection regulations necessitating the use of synthetic data.

\subsection{Reproducibility of Results}

We define reproducibility as a factor of the use of publicly available datasets and the reporting of publicly accessible source code of the research. We note that, despite being a crucial piece of information, reproducibility is often not emphasised, and only $24.56$\% of all included publications were reproducible according to our definition. The majority of the papers ($64.91$\%) use publicly available real-world health datasets, whereas about $11.22$\% used paid real-world health datasets for the creation of synthetic data.  However, only $25.50$\% of the publications give details about their code along with a link to the code repository, which affects the overall reproducibility. Reproducibility is essential for ensuring the reliability and impact of scientific research. Particularly in healthcare, where decisions can directly affect patient outcomes, reproducibility helps prevent errors, biases, and misleading conclusions. However, we found that it is an often overlooked aspect in most publications dealing with synthetic data.

In response to these results, we propose the following set of reporting guidelines on the generation and evaluation of synthetic data, which are henceforth described in Section \ref{Section:guidelines}.

\section{Evaluation Guidelines for Synthetic Data} \label{Section:guidelines}

\begin{enumerate}
    \item \textbf{Standardised Evaluation of Synthetic data:} We found that synthetic data are sometimes used without a thorough assessment. When there is an assessment, we found that not only is there no consensus on the evaluation methods, but the chosen evaluation metrics are often inconsistently applied. This makes an operational assessment of the entire process unreliable, thereby making it difficult to track state-of-the-art advancements and creating barriers to trust and the adoption of synthetic data. \textit{For example}, using Mean Squared Error (MSE) metric as a measure of distortion to assess the validity of a synthetically generated waveform signal is appropriate, but the validation needs to consider the particular features of the process. Two synthetic data generation methods with comparable MSE performances can still yield qualitatively very different signal features. Recent works, such as the ‘Scorecard for Synthetic Medical Data Evaluation’\cite{zamzmi2025scorecard} have also been proposing structured frameworks (e.g., the ‘7 Cs’) that emphasise multi-dimensional assessment of synthetic data quality, further highlighting the need for comprehensive and standardised evaluation approaches.

    \item \textbf{Better Reporting of Dataset Characteristics:} Poor reporting of dataset details is a cause for concern since the type of data (such as continuous vs categorical) and their distribution significantly impact the quality of the generated synthetic data. Furthermore, potential biases may propagate in the synthetic data. We recommend improved reporting of dataset characteristics. \textit{For example}, in developing a novel method, one would expect to assess its performance using real data or previously validated synthetic data. Using synthetic data that has not been validated prior to the assessment will not provide robust evidence towards the validity of the method. Moreover, any claims about the validity of the synthetic data based on the performance of the proposed method are inherently inconclusive, as they incur a circular reference problem that compromises the generated evidence.

    \item \textbf{Prioritisation of  Reproducibility of Results:} 
    We emphasise the importance of clearly described evaluation metrics used, normalisation and aggregation of real data that were used as training datasets, and the entire experimental setup for the generation of synthetic data, including the chosen hyperparameters and the source code, where possible. Reproducibility allows other researchers to validate and verify the claims of a study, and stakeholders, including clinicians and patients to develop trust in synthetic health data.

\end{enumerate}

\section{Conclusion}

The potential of synthetic data to transform health AI research is immense, offering promising solutions to data scarcity, improved privacy preservation, accelerated clinical trials, mitigation of representation biases, and more flexible model development. Realising this potential depends on rigorous and reliable evaluation frameworks that ensure synthetic data is fit for purpose, clinically meaningful, and scientifically robust. Alongside this, transparent reporting of synthetic datasets and reproducibility of experimental results remain essential requirements for trustworthy scientific progress. 

A key contribution of this review is the systematic consolidation of generation and evaluation methods into structured taxonomies for tabular data. We highlight an ongoing need for clearer definitions, improved standardisation, and more consistent reporting practices, particularly in safety-critical healthcare applications. Meaningful evaluation of synthetic health data also requires close engagement with domain expertise, underscoring the importance of interdisciplinary evaluation practices. 

At the same time, the development and deployment of synthetic health data are increasingly shaped by ethical and regulatory expectations. Frameworks such as the European Union’s Artificial Intelligence Act\cite{EU_AI_Act}, guidance from regulatory bodies such as the UK Medicines and Healthcare products Regulatory Agency (MHRA)\cite{MHRA}, and broader U.S. governance frameworks including HIPAA\cite{HIPAA} privacy requirements and the NIST AI Risk Management Framework\cite{NIST_AI_Framework}, emphasise requirements around transparency, accountability, robustness, and risk management in AI systems. In practice, these expectations imply that synthetic datasets should be accompanied by clear documentation of generation processes, validation protocols, and intended use cases to support regulatory compliance and responsible deployment. Beyond formal regulation, institutional governance mechanisms and research oversight processes will also play a critical role in ensuring appropriate use.

Synthetic data presents opportunities for positive societal impact, including improving representativeness in healthcare datasets, mitigating health inequalities, enabling broader data access and democratisation, and reducing risks associated with sharing sensitive patient information. However, risks such as residual re-identification potential, the propagation of latent biases, and overestimation of data fidelity highlight the importance of careful evaluation and governance. Ensuring that evaluation frameworks reflect both technical performance and these broader ethical and societal dimensions will be critical for responsible adoption.

By adopting structured evaluation practices, advancing standardised reporting, and aligning evaluation approaches with emerging regulatory and governance expectations, the health AI community can ensure that synthetic data is developed and used in a manner that is both effective and responsible, ultimately supporting innovation and improving patient outcomes.


\bibliographystyle{unsrt}
\bibliography{software}


\appendix

\section{Methodology}

In this section of the Appendix, we provide additional details about the methodology used to carry out the systematic review.

\subsection{Search Queries} \label{appendix:searchqueries}

Table \ref{tab:queries} lists the search queries used across five databases, and the number of relevant results obtained from each. It should be noted that additional filtering criteria were set on these databases, including the date of publication range (2014-2026) and the language of the publication (English).

\begin{table}[htbp]
\centering
\small
\begin{tabular}{p{1.75cm} p{10cm} p{2cm}}
\toprule
\textbf{Database} & \textbf{Query} & \textbf{No. of Results} \\
\midrule

Scopus &
(synthe* OR augment*) AND generat* AND (time-series OR time* OR temporal*) AND (tabular OR record*) AND (patient* OR medic* OR health* OR clinic* OR ehr*) on title-abstract-keywords
& 487 \\

Web of Science &
(synthe* OR augment*) AND generat* AND (time-series OR time* OR temporal*) AND (tabular OR record*) AND (patient* OR medic* OR health* OR clinic* OR ehr*) (Topic)
& 1173 \\

PubMed &
(synthe* [Title/Abstract] OR augment* [Title/Abstract]) AND generat*[Title/Abstract] AND (time-series[Title/Abstract] OR time*[Title/Abstract] OR temporal*[Title/Abstract]) AND (tabular[Title/Abstract] OR record*[Title/Abstract]) AND (patient*[Title/Abstract] OR medic*[Title/Abstract] OR health*[Title/Abstract] OR clinic*[Title/Abstract])
& 237 \\

IEEE Xplore &
("All Metadata":synthe* OR "All Metadata":augment*) AND ("All Metadata":time* OR "All Metadata":temporal*) AND ("All Metadata":tabular OR "All Metadata":record*) AND ("All Metadata":patient* OR "All Metadata":medic* OR "All Metadata":health* OR "All Metadata":clinic* OR "All Metadata":ehr*) AND ("All Metadata":generat*)
& 158 \\

ACM &
[Abstract: synthe*] AND ([Abstract: time*] OR [Abstract: temporal*]) AND ([Abstract: patient*] OR [Abstract: medic*] OR [Abstract: health*] OR [Abstract: clinic*] OR [Abstract: ehr*])
& 12 \\

\bottomrule
\end{tabular}
\caption{Search Queries and the number of Results obtained from each database}
\label{tab:queries}
\end{table}

\subsection{Selection Process}

Based on the search strategy discussed in Section \ref{SearchStrategy}, 2067 publications were obtained from five search engines. This included 487  publications from Scopus, 1173 from the
Web of Science, 237 from PubMed, 158 from IEEE Xplore, and 12 from ACM. Since
many results were duplicated across databases, we carried out a deduplication process based on the DOIs of publications. As a result, 992 results were excluded, and 1077 publications remained for the first stage of assessment.

\paragraph{First Screening}

The 1077 publications obtained from search queries after deduplication, were divided among five reviewers, with approximately 215 publications per reviewer. Each reviewer labelled the publications assigned to them Yes/No, signifying whether they thought the publication should be included or excluded, along with the rationale for their decision. This gave us 195 publications labelled 'Yes', and 882 publications labelled 'No'. From the 'No' set, we again discarded some publications, the most common reasons for their exclusion being that the research was not in the human health domain (n=336) or that the paper was a systematic review itself (n=163). Next, we created a smaller subset of approximately 10\% of the remaining No-labelled publications and added it to all publications in the Yes-labelled set. This combined set of 233 publications was used to perform a 'spot check' of labels: Publications were grouped by their first-stage reviewers and divided among the rest of the reviewers for a second round of reviewing. The first-stage reviewer's decision to include or exclude any particular publication was preserved but kept hidden from the view of the second-stage reviewers, to ensure that the reviewers' cognitive biases do not creep in during the labelling process, and that every publication gets assigned the correct label, irrespective of who it was reviewed by in either of the two reviewing stages. 

\paragraph{Second Screening}

As with the first stage, each reviewer in the second stage provided a 'Yes' or a 'No' label to each publication in their set. No reviewer got to review the same publication in the second stage which they had already reviewed in the first stage. At the end of this exercise, any discrepancies (cases where the labels given by the first-stage reviewer and the second-stage reviewer did not match) were duly noted. Then, a round of discrepancy check was carried out, where all reviewers looked at all discrepancies and provided feedback as to which of the two labels they agreed with. After a discussion on each occurrence of conflicting labels and resolving all discrepancies, we got the final labels for each publication. This resulted in a set of $158$ publications for which the consensus of the reviewers was to include them in this systematic review after an in-depth analysis. For each of the $158$ publications, an in-depth analysis was carried out. An additional $24$ publications were excluded from the study upon full-text analysis, based on their relevance to this research. 

\subsection{Data Items}

Data was collected for the final $134$ publications for 18 attributes which included: \textbf{(i.)} details about the publication including DOI, authors' names, title and year of publication and the details about the venue (journal/conference), \textbf{(ii.)} details specific to the generation of synthetic data such as the method (eg. WGAN-GP, Graph VAEs), its category (mechanistic/probabilistic), and purpose (eg. privacy preservation, clinical trial simulation), \textbf{(iii.)} details about the evaluation methods used, which includes the type of Quantitative evaluation (ML-based, Statistical or a combination), the type of Qualitative evaluation (for eg. Visualisation), the name of the method (eg. Jensen-Shannon divergence, Wasserstein distance) and the specific evaluation metrics used, \textbf{(iv.)} details about the dataset used in synthetic data generation, including the dataset name, size, institution and country of origin, and cost of dataset access, and \textbf{(v.)} details pertaining to the reproducibility of results including whether the dataset is openly accessible and if the source code has been made available. A complete list of all data items against which data was captured is available in Table \ref{tab:dataItems}.

\begin{table}[htbp]
\centering
\small
\renewcommand{\arraystretch}{1.2}
\begin{tabular}{p{3cm} p{11cm}}
\toprule
\textbf{Category} & \textbf{Data Items} \\
\midrule

Publication Details &
DOI, Authors, Title, Publication Year, Journal/Conference Title \\

Synthetic Data Generation &
Category of Generation Model (Mechanistic, Probabilistic), Name of Generation Method used, Purpose of Generation of synthetic data, Disease/Disorder focused on, ICD-9 code \\

Synthetic Data Evaluation &
Category of Quantitative Evaluation Method (ML-based, Statistical, Both, None), Category of Qualitative Evaluation Method (Visualisation, Others, None), Name of Evaluation Method used \\

Training Dataset Characteristics &
Name, Size, Institution of Origin, Country of Origin, Visibility (Public/Private), Cost of Dataset Access \\

Source Code &
Link to Source Code Repository \\

\bottomrule
\end{tabular}
\caption{Data extraction categories and items}
\label{tab:dataItems}
\end{table}

\section{Additional Results}

Table \ref{tab:tools} contains a non-exhaustive list of existing tools and libraries used for synthetic data generation and evaluation, based on publications included in this review.

\begin{table}[]
\small
\begin{tabular}{@{}ll@{}}
\toprule
\textbf{Tool}      & \textbf{Link}                            \\ \midrule

SynthCity & https://pypi.org/project/synthcity/      \\
PyTrial  & https://pypi.org/project/PyTrial/    \\
OSIM      & https://github.com/OHDSI/OSIM-v5         \\
GANerAid  & https://pypi.org/project/GANerAid/ \\
synthetic\_data &
  https://github.com/TheRensselaerIDEA/synthetic\_data \\
HealthGen & https://github.com/simonbing/healthgen \\ 

\bottomrule
\end{tabular}
\caption{Tools for synthetic data generation and evaluation}
\label{tab:tools}
\end{table}

\section{List of papers reviewed} \label{appendix:sheet1}
An abridged version of the papers reviewed and their characteristics is provided in Table \ref{tab:my-table3}



\begin{small} 
\begin{longtable}
[l]{p{2cm}p{2.5cm}p{3.5cm}p{3.5cm}p{2cm}}

\hline
\multicolumn{1}{c}{\textbf{Publication}} &
  \multicolumn{1}{c}{\textbf{Generation Method}} &
  \multicolumn{1}{c}{\textbf{Evaluation Metrics}} &
  \multicolumn{1}{c}{\textbf{Datasets Used}} &
  \multicolumn{1}{c}{\textbf{Purpose}} \\ \hline
  
\textbf{Jatoi et al, 2018} \cite{Jatoi2018BrainSL} &
  Custom method &
  Negative variational free energy, Localization error &
  Statistical Parametric Mapping - SPM12 software &
  Predictive modelling \\
\textbf{Farago et al, 2021} \cite{articleFarago} &
  Autoregressive modeling, Markov chain, RNN &
  Morphology, Mean, Variance, Autocorrelation, Power Spectral Density (PSD), Probability Distribution &
  Custom dataset &
  Signal quality analysis \\
\textbf{Kuo et al, 2022} \cite{kuo2022towards} &
  SAGAN &
  Accuracy, Standard Deviation &
  Custom dataset &
  Improve personalisation of prediction \\
\textbf{Rafiei et al, 2023} \cite{rafiei2023improving} &
  CTGAN and SMOTE &
  AUC, AUROC, Sensitivity, Specificity, PPV, NPV, Bhattacharyya Distance &
  North Carolina Health System electronic medical record (EMR) &
  Fluid overload \\
\textbf{DeOliveira et al, 2022} \cite{deoliveira2022har} &
  CTGAN and HAR-CTGAN &
  Weighted Average F1-score, Ambiguity score &
  ExtraSensory dataset &
  Generating discreet synthetic data \\
\textbf{Das et al, 2023} \cite{das2023twin} &
  VAE &
  Dimension-Wise Probability, Bernoulli Success Probability, Counterfactual Digital Twin Evaluation, Presence Disclosure, Attribute Disclosure, Nearest Neighbor Adversarial Accuracy Risk &
  Phase III breast cancer clinical trial (NCT00174655), Small Cell Lung Carcinoma clinical trial dataset (NCT01439568), &
  Clinical trials \\
\textbf{Bing et al, 2022} \cite{bing2022conditional} &
  VAE &
  KNN &
  MIMIC-III &
  Mitigating representation bias \\
\textbf{Bhanot et al, 2021} \cite{bhanot2021quantifying} &
  HealthGAN &
  Root Mean Square Error (RMSE), Pearson’s Correlation Coefficient, Directional Symmetry, Short Time-Series Distance &
  American Time Use Survey (ATUS), Medical claims Autism Spectrum Disorder (ASD) &
  Privacy preservation, maintaining utility \\
\textbf{El Emam et al, 2020} \cite{el2020evaluating} &
  Conditional trees &
  Matching real with synthetic samples &
  Washington State Inpatient Database (SID) and Canadian COVID-19 case dataset &
  Privacy preservation \\
\textbf{Lu et al, 2023} \cite{lu2023explainable} &
  SMOTE &
  Precision, Recall, F1-score, Geometric mean, Area under the curve of the receiver operating characteristic curve (AUROC), Area under the precision-recall curve (AUPRC) &
  Taipei Medical University Hospital and Wan Fang Hospital (derivation), Taipei Medical University Shuang Ho Hospital (validation) &
  Predictive modelling \\
\textbf{Solinski et al, 2016} \cite{solinski2016modeling} &
  Detrended fluctuation analysis (DFA) &
  Shannon Entropy, Poincaré Plots, Multiscale Multifractal Analysis &
  Holter electrocardiogram (ECG) database and Complete electroencephalogram (EEG) recordings &
  NA \\
\textbf{Bhanot et al, 2021} \cite{bhanot2021problem} &
  HealthGAN &
  Log Disparity, Time-Series Disparity &
  MIMIC-III and Average Sleep Time of Americans (ATUS) &
  Fairness \\
\textbf{Shaked et al, 2016} \cite{shaked2016publishing} &
  Markovian Model &
  Mean Similarity, Intersection &
  MIMIC-III &
  Privacy preservation \\
\textbf{Bahador et al, 2021} \cite{bahador2021morphology} &
  DE-NLPCA &
  Accuracy &
  Activities of daily living (ADL) dataset, and EEG / ECG dataset from Northern Ostrobothnia Hospital &
  Predictive modelling \\
\textbf{Weng et al, 2024} \cite{weng2024joint} &
  MVIIL-GAN &
  Missing Values Reconstruction Error &
  MIMIC-IV &
  Dataset balancing \\
\textbf{Bhanot et al, 2022}\cite{bhanot2022investigating} &
  Bootstrapping, Random Permutation and HealthGAN &
  Root Mean Square Error (RMSE), Pearson’s Correlation Coefficient, Short Time-Series Distance (STS), Directional Symmetry (DS) &
  American Time Use Survey (ATUS) dataset and Autism Spectrum Disorder (ASD) claims dataset. &
  Addressing Data unavailability, Privacy preservation \\
\textbf{Nikolentzos et al, 2023} \cite{nikolentzos2023synthetic} &
  Variational Graph Autoencoder (VGAE) &
  Weisfeiler-Lehman Subtree (WL) Graph Kernel, Shortest Path (SP) Graph Kernel, Graph Kernel-Maximum Mean Discrepancy (GK-MMD),  Pearson Correlation Coefficient &
  MIMIC-IV &
  Privacy preservation \\
\textbf{Ricci et al, 2019} \cite{ricci2019generation} &
  Custom method &
  Poisson Sequence, Correlated Heartbeatlike Sequence, Hénon Map Sequence &
  Electronic oscillator sequence, Heartbeat sequence, and Neural sequence &
  NA \\
\textbf{Hodgkiss et al, 2021} \cite{hodgkiss2021new} &
  Custom method &
  AUC &
  Normal Sinus Rhythm Dataset &
  Cybersecurity \\
\textbf{Chen et al, 2022} \cite{chen2022epileptic} &
  CTGAN &
  Classifier &
  CHB-MIT EEG dataset &
  Dataset balancing \\
\textbf{Xu et al, 2017} \cite{xu2016patient} &
  Custom method &
  Markov Chain, Vector Auto-Regressive Model, Continuous-Time Markov Chain, Logistic Regression, Hawkes Processes, Modulated Poisson Processes, Self-Correcting Process &
  MIMIC-II &
  Dataset balancing \\
\textbf{Mohammed et al, 2017} \cite{mohammed2017toward} &
  ARMA &
  F1-score, AUC, NOP &
  Custom dataset &
  Addressing Data unavailability \\
\textbf{Wang et al, 2023} \cite{wang2023afe} &
  AFE-GAN (Atrial Fibrillation-like ECG GAN) &
  Two of the four winning atrial fibrillation detectors from the 2017 PhysioNet Challenge - Hong detector, Datta detector &
  training set from the 2017 PhysioNet Challenge &
  Addressing Data unavailability \\
\textbf{Imtiaz et al, 2021} \cite{imtiaz2021synthetic} &
  BGAN (Boundary-seeking GAN) &
  Visualisation only &
  Custom dataset - from Fitbit Charge 2 HR smartwatches &
  Privacy preservation \\
\textbf{Ashrafi et al, 2023} \cite{ashrafi2023protect} &
  simpleGAN, medGAN, DoppelGANger, DPGAN, and PPGAN &
  F1-score, Precision, Recall, Root Mean Square Error (RMSE), AUC, Attacker Advantage &
  (Patient interactions with a tablet game (PflegeTab) ) &
  Privacy preservation \\
\textbf{Rafiei et al, 2024} \cite{rafiei2024improving} &
  SMOTE, CTGAN &
  Jensen-Shannon Divergence (JSD), Bhattacharyya Distance, Mann-Whitney U Test, Benjamini-Hochberg (BH) procedure &
  Custom dataset &
  Predictive modelling \\
\textbf{Agliari et al, 2020} \cite{agliari2020analysis} &
  Custom method &
  Power Spectrum Density (PSD) &
  Custom dataset &
  Method evaluation \\
\textbf{Li et al, 2023} \cite{li2023generating} &
  EHR-M-GAN and EHR-M-GANconditional &
  Maximum Mean Discrepancy (MMD), Dimension-Wise Probability, Discriminative Score, Patient Trajectories, Pearson pairwise correlations, Autocorrelation function, Membership Inference Attack, Differential Privacy &
  MIMIC-III, eICU and HiRID &
  Addressing Data unavailability, Privacy preservation \\
\textbf{Zhang et al, 2022} \cite{zhang2022keeping} &
  LS-EHR &
  Jensen–Shannon Divergence (JSD), AUC &
  Custom datasets (two) &
  NA \\
\textbf{Oh et al, 2023} \cite{oh2023validating} &
  Monte Carlo simulations &
  Relative Bias, Confidence Limit Ratios (CLRs), Mean Square Error (MSE) &
  Custom dataset (South Korea's patients’ healthcare resource utilization database) &
  Checking bias \\
\textbf{Burrello et al, 2022} \cite{burrello2022improving} &
  DA techniques and DL HR algorithms &
  Visualisation only &
  PPGDalia &
  Health monitoring \\
\textbf{Mendez et al, 2022} \cite{mendez2022emg} &
  GAN &
  Welch's Test &
  NA &
  Health monitoring and Data privacy \\
\textbf{Yang et al, 2022} \cite{yang2022discovery} &
  Physics-based models &
  DFA &
  PhysioNet &
  Predictive modelling \\
\textbf{Vulpe-Grigorasi et al, 2022} \cite{vulpe2022gan} &
  GAN &
  RMSSD, SDNN &
  PhysioNet &
  Increased diagnosis accuracy \\
\textbf{Shamsuddin et al, 2018} \cite{shamsuddin2018virtual} &
  Virtual Patient Model &
  NB, SVM and TB &
  ARem and EEG &
  Addressing Data unavailability \\
\textbf{Balasubramanian et al, 2016} \cite{balasubramanian2016discovering} &
  MDMs (Multidimensional Motifs) &
  Graph Clustering Method &
  Electromagnetic Articulography and Motion Capture and Muscle Activity &
  Personalised diagnosis and therapy \\
\textbf{Tessier-Larivière et al, 2021} \cite{tessier2021pns} &
  PNS-GAN &
  Power Spectral Density, Euclidean distance &
  BIOS-IT3 Dataset &
  Data Augmentation \\
\textbf{Vandendriessche et al, 2017} \cite{vandendriessche2017framework} &
  MSE &
  Classifier &
  MIMIC-III (heart and sepsis) &
  Predictive modelling \\
\textbf{Tasnim et al, 2022} \cite{tasnim2022approach} &
  SMOTE &
  Classifier &
  BCIAUT-P300 &
  Addressing Health inequality \\
\textbf{Wang et al, 2023} \cite{wang2023enhancing} &
  SMOTE and WCGAN-GP &
  Classifier &
  ImmPort (Immunology Database and Analysis Portal) data &
  Enhancing health clinical data \\
\textbf{García-Vicente et al, 2023} \cite{garcia2023evaluation} &
  SMOTE, CTGAN, TVAE &
  LASSO, SVM, KNN, DT &
  Norwegian Centre for E-health Research &
  Data quality enhancement \\
\textbf{Kuo et al, 2022} \cite{kuo2022health} &
  GAN &
  Classifier &
  MIMIC-III and EuResist23 &
  Data quality enhancement, Privacy preservation \\
\textbf{Yoon et al, 2023} \cite{yoon2023ehr} &
  Sequential encoder-decoder methods and GAN &
  Classifier &
  MIMIC-III &
  Privacy preservation \\
\textbf{Mitra et al, 2023} \cite{mitra2023cardiosim} &
  CardioSim PC-based system &
  Classifier &
  mitdb &
  Data quality enhancement \\
\textbf{Khan et al, 2016} \cite{khan2016using} &
  NA &
  MEWMA &
  accelerometer &
  Health and wellbeing  assessment \\
\textbf{Roonizi et al, 2015} \cite{roonizi2015comparison} &
  SHVR &
  Mean Square Error (MSE), Wilcoxon Rank Test &
  synthetic ECG &
  Predictive modelling \\
\textbf{Konak et al, 2023} \cite{konak2023overcoming} &
  TimeGAN and Animations &
  MMD &
  PAMAP2 and SONAR-LAB &
  Addressing Data unavailability \\
\textbf{Tan et al, 2023} \cite{tan2023tabular} &
  TabGAN and SMOTE &
  Jensen–Shannon Divergence (JSD), Wasserstein Distance (WD), Diff Corr &
  SUPPORT and METABRIC &
  Predictive modelling \\
\textbf{Liu et al, 2020} \cite{liu2020estimating} &
  Deep Sequential Weighting (DSW) &
  LR, RF, KNN, PSM, CFR, CF, BART &
  Custom dataset and MIMIC-III &
  Predicitve modelling \\
\textbf{Atmadja et al, 2022} \cite{atmadja2022generated} &
  GAN &
  CNN &
  MIT-BIH &
  Predictive modelling \\
\textbf{Hashemi et al, 2023} \cite{hashemi2023time} &
  GAN &
  PCA, t-SNE, pairwise correlations,RNN &
  Sins, MIMIC-VI &
  Privacy preservation \\
\textbf{Wang et al, 2022} \cite{Wang2022UsingAO} &
  Markov Jump Process &
  Domain Expert Review &
  NA &
  Predictive modelling \\
\textbf{Huerta et al, 2021} \cite{Huerta2021ECGQA} &
  Standard Data Augmentation Transformations &
  McNemar test &
  PhysioNet/CinC Challenge 2017 database &
  Addressing Data
unavailability \\
\textbf{Ayilara et al, 2023} \cite{Ayilara2023GeneratingSD} &
  OSIM2 and ModOSIM &
  Concordance Correlation Coefficient &
  Population Research Data Repository (PRDR) &
  Method evaluation \\
\textbf{Haleem et al, 2023} \cite{Haleem2023DeepLearningDrivenTF} &
  TC-Multi GAN and Document Sequence Generator &
  Wasserstein Distance, Kolmogorov-Smirnov Test, Jensen-Shannon Distance, Distance Pairwise Correlation, Sample Kernel Density Estimations &
  GATEKEEPER EU project &
  Addressing Synthetic Data feasibility \\
\textbf{Zama et al, 2023} \cite{Zama2023ECGSV} &
  Diffusion-based model &
  Dynamic Time Warping, Maximum Mean Discrepancy &
  PTB-XL &
  Privacy preservation \\
\textbf{Mosquera et al, 2023} \cite{Mosquera2023AMF} &
  RNN with LTSM and GRU &
  Hellinger's Distance, Cox Regression Hazard Ratios &
  Alberta Health's administrative dataset &
  Addressing Synthetic Data feasibility \\
\textbf{Huang et al, 2019} \cite{Huang2019EvaluatingGA} &
  Delete, update, switch operations &
  Pairwise Similarity Score &
  Rochester epidemiology project &
  Predictive modelling \\
\textbf{Dahal et al, 2023} \cite{Dahal2022AHG} &
  EC–WCGAN &
  Precision, Recall, F1-score &
  AHADB, VFDB and CUDB &
  Dataset balancing \\
\textbf{Setiawan et al, 2022} \cite{Setiawan2022ADL} &
  SMOTE &
  Accuracy, Sensitivity, Specificity, ROC, Cross-Validation, MSE, MAE &
  PhysioNet Apnea-ECG database (PAED) &
  Dataset balancing \\
\textbf{Jurdana et al, 2023} \cite{Jurdana2023MethodFA} &
  Custom method &
  MSE &
  EEG, Royal Brisbane &
  Predictive modelling \\
  
\textbf{Yang et al, 2023} 
\cite{Yang2023TSGANTG} &
  TS-GAN &
  LSTM-based Discriminator, Discriminator loss, Maximum Mean Discrepancy, Principal Component Analysis, t-SNE, Sequence diagrams, Accuracy &
  ECG\_200, NonInvasiveFatalECG\_Thorax1, and mHealth &
  Data augmentation \\
\textbf{Mahey et al, 2023} \cite{Mahey2023GenerativeAN} &
  Simulation &
  Channel by Channel Covariance, EOG, 1/f Function, Spatial Covariance &
  NA &
  Addressing Data unavailability \\
\textbf{Velasco et al, 2018} \cite{Velasco2018CombiningDA} &
  Evolutionary algorithm &
  Wilcoxon Rank Sum Test (Mann Whitney Wilcoxon) (MWW) &
  Principe de Asturias Hospital &
  Addressing Data unavailability \\
\textbf{Xiong et al, 2022} \cite{Xiong2022EnhancingTD} &
  Custom method &
  Mean Square Error (MSE), Average Standard Deviation, Frequency Distribution &
  PhysioNet 2017 challenge dataset &
  Dataset balancing \\
\textbf{Krenmayr et al, 2022} \cite{Krenmayr2022GANerAidRS} &
  GAN with bi-LSTM &
  Euclidean Distance, Wasserstein Distance &
  NA &
  Addressing Data unavailability \\
\textbf{Zaballa et al, 2023} \cite{Zaballa2022LearningTP} &
  Probabilistic generative model (HMM and EM) &
  Average Log Likelihood &
  NA &
  Predictive modelling \\
\textbf{Wendland et al, 2022} \cite{Wendland2021GenerationOR} &
  Multimodal Neural Ordinary Differential Equations &
  Jensen-Shannon Divergence &
  PPMI (Parkinson), and NACC (Alzheimer) &
  Predictive modelling \\
\textbf{Kaleli et al, 2023} \cite{Kaleli2023GenerationOS} &
  GAN with CNN and Transformer &
  Percent Root Mean Square Difference (PRD), Root Mean Square Error (RMSE), Frechet Distance (FD) &
  MIT-BIH dataset &
  Privacy preservation, Predictive modelling \\
\textbf{Sun et al, 2021} \cite{Sun2021GeneratingLS} &
  Longitudinal GAN &
  AUROC, AUPCR, AUC &
  Cerner Health Facts database &
  Predictive modelling, Privacy Preservation \\
\textbf{Shilaskar et al, 2017} \cite{Shilaskar2017MedicalDS} &
  Resampling, modified Particle Swarm Optimization &
  Accuracy, Precision, Recall, Sensitivity, F1-score &
  Vani Dataset, Thyroid Dataset, PdA, Cleveland, Audiology, SVD, Vertigo &
  Predictive modelling \\
\textbf{Foomani et al, 2022} \cite{Foomani2021SynthesizingTW} &
  GAN &
  Jensen-Shannon Divergence, AUC &
  EMR data from Vascular Centers, Milwaukee, WI &
  Predictive modelling \\
\textbf{Peng et al, 2022} \cite{Peng2021ImpulseDM} &
  Gaussian Kernels &
  Root Mean Square Error (RMSE) &
  NA &
  Predictive modelling \\
\textbf{Geng et al, 2021} \cite{Geng2021AuxiliaryCG} &
  GAN &
  AUC, F1-score &
  NA &
  Predictive modelling \\
\textbf{Mivalt et al, 2022} \cite{Mivalt2022DeepGN} &
  GAN &
  Cohen's Kappa, F1-score &
  Multicenter Intracranial EEG Dataset &
  NA \\
\textbf{Pascual et al, 2021} \cite{Pascual2020EpilepsyGANSE} &
  GAN &
  Cosine Similarity, Recall &
  EPILEPSIAE &
  Privacy preservation \\
\textbf{Lu et al, 2023} \cite{Lu2022MultiLabelCT} &
  GAN &
  Jensen-Shannon Divergence, Normalised Distance, AUC, F1-score &
  MIMIC-III and MIMIC-IV &
  Privacy preservation \\
\textbf{Ikuma et al, 2016}\cite{Ikuma2016SyntheticMK} &
  Karhunen-Loeve transformation, Time-series model perturbations &
  Correlation analysis relative vibration power represented by a synthetic waveform &
  NA &
  Predictive modelling \\
\textbf{Zhang et al, 2021} \cite{Zhang2020SynTEGAF} &
  GAN &
  Diagnosis forecast analysis, Kolmogorov-Smirnov Test &
  Synthetic Derivative at Vanderbilt University Medical Center &
  Privacy preservation \\
\textbf{Baowaly et al, 2019} \cite{Baowaly2019RealisticDS} &
  GAN &
  Dimension-Wise Average, Kolmogorov-Smirnov (KS) Test, Association Rule Mining &
  MIMIC-III and NHIRD &
  Predictive modelling \\
\textbf{Abdelghaffar et al, 2022} \cite{Abdelghaffar2022GenerativeAN} &
  GAN &
  Relative Entropy, Accuracy &
  Wadsworth BCI Dataset from the BCI competition III &
  Predictive modelling \\
\textbf{Al-Saad et al, 2021} \cite{AlSaad2021PrivacyVO} &
  DPGAN &
  Dimension-Wise Average, AUROC, Area under the Precision-Recall Curve, Accuracy &
  Arizona State’s Kinesiology Department &
  Predictive modelling, Privacy preservation \\
\textbf{Yu et al, 2023} \cite{Yu2023RefineES} &
  GAN &
  AUC &
  CHB-MIT dataset &
  Predictive modelling \\
\textbf{Cenek et al, 2020} \cite{Cenek2020CIRCADASA} &
  Frequency Domain Model &
  NA &
  CIRCADA-S &
  Predictive modelling \\
\textbf{Yadav et al, 2023} \cite{Yadav2023QualitativeAQ} &
  GAN &
  Mean Absolute Error (MAE), MRLE, PCA, t-SNE &
  UNIMIB &
  Predictive modelling \\
\textbf{Yu et al, 2024} \cite{Yu2023ElectrophysiologicalBI} &
  Temporal Convolutional Network &
  Dipole Localization Error (DLE), Normalized Hamming Distance, Sensitivity, Specificity, False Detection Rate, F1-score, Pearson Correlation &
  NA &
  Predictive modelling \\
\textbf{Walonoski et al, 2018} \cite{Walonoski2017SyntheaAA} &
  Markovian model (PADARSER) &
  Prevalence Difference Error &
  Multiple &
  Predictive modelling \\
\textbf{Kulpa et al, 2022} \cite{Kulpa2022ATF} &
  Autoregressive model, Markov chain, RNN &
  Power Spectral Density (PSD) &
  MIT-BIH NSTDB &
  Predictive modelling \\
\textbf{Kan et al, 2021} \cite{Kan2021EnhancingME} &
  GAN &
  Average Error Rate &
  Temple University Hospital Abnormal EEG Corpus &
  Predictive modelling \\
\textbf{Naseer et al, 2023} \cite{Naseer2023ScoEHRGS} &
  Continuous-Time Diffusion Models &
  Dimension-wise distribution, Pairwise Correlation difference, Log-cluster, Synthetic ranking agreement, Membership Inference Attack, Blinded Clinician Evaluation, Domain Expert Review &
  MIMIC-III and ED-EHR datasets &
  NA \\
\textbf{Qian et al, 2024} \cite{Qian2023SyntheticDF} &
  DPGAN, PATEGAN), ADSGAN &
  Fidelity (Alpha-Precision), Diversity (Beta-Recall), Authenticity, Wasserstein distance, Jensen-Shannon distance, Inverse Kullback-Leibler divergence, Chi-Squared Tes, Kolmogorov-Smirnov test, k-anonymity, DOMIAS AUC &
  Ever-smokers in UK Biobank Database &
  Privacy preservation \\
\textbf{Lu et al, 2023} \cite{lu2023explainable} &
  SMOTE &
  Decision Tree, Random Forest, Logistic Regression, Extreme Gradient Boosting, Support Vector Machines &
  Custom data &
  Dataset balancing \\
\textbf{Sawant et al, 2021} \cite{Sawant2021AutomatedDO} &
  SMOTE &
  Sensitivity, Specificity, Overall score (Average of Sensitivity and Specificity) &
  PhysioNet / CinC challenge 2016, and PASCAL &
  Dataset balancing \\
\textbf{Akter et al, 2021} \cite{akter2021towards} &
  SMOTENC (SMOTE variant) &
  Classifier &
  Quantitative Checklist for Autism in Toddlers-10 (Q-CHAT-10), and Autism Spectrum Quotient-10 (AQ-10) &
  Predictive modelling \\
\textbf{Katekarn et al, 2023} \cite{katekarn2023studying} &
  Custom method &
  Participant satisfaction questionnaire and SPSS &
  Custom dataset &
  Method evaluation \\
\textbf{Vemuri et al, 2016} \cite{vemuri2015interoperative} &
  Custom method &
  Custom evaluation metrics (Measurement of Uncertainty, Measuring Uncertainty in Endoscope Tip) &
  NA &
  Predictive modelling \\
\textbf{Valdano et al, 2015} \cite{Valdano2014PredictingER} &
  Custom method &
  Visualisation only &
  NA &
  Disease modelling \\
\textbf{Covioli et al, 2023} \cite{Covioli2023WorkflowCO} &
  Simulation &
  Visualisation only &
  MIMIC-III and MIMIC-III waveform matched dataset &
  Method evaluation \\
\textbf{Tomek et al, 2016}\cite{Tomek2016CcoffinnAW} &
  Cellular automata &
  Median, p-value &
  Custom dataset &
  Method evaluation \\
\textbf{Squires et al, 2022} \cite{Squires2022ANG} &
  Python packages (random and fake) &
  Custom evaluation metrics &
  Custom dataset &
  Addressing Data unavailability \\
\textbf{Oliveira et al, 2023} \cite{oliveira2023tabular} &
  CGAN and CT-GAN &
  False Positives, False Negatives, Accuracy, Specificity, Sensitivity, AUC &
  PARK Facial Mimic &
  Data Augmentation \\

  \textbf{Budu et al, 2026} \cite{budu2025tempehr} &
  VAE &
  UMAP, DTW, Membership Inference Attack &
  MIMIC-IV, MIMIC-CHF &
  Addressing Data unavailability \\
  \textbf{Liu et al, 2025} \cite{liu2025augmenting} &
  Sequential decision trees, Bayesian networks, Conditional GAN, Tabular VAE &
  AUC-ROC &
  Custom dataset  &
  Dataset balancing \\
  \textbf{Kulyabin et al, 2025} \cite{kulyabin2025synthetic} &
  Conditional GAN &
  Accuracy, Precision, Recall, F1-score, AUC, Domain Expert review &
  Electroretinogram Raw Waveforms &
  Dataset balancing \\
  \textbf{Simone et al, 2025} \cite{simone2025ecg} &
  ECGAN &
  Inception Score, Frechet Inception Distance, Maximum-Mean-Discrepancy, Wasserstein distance, Accuracy, Sensitivity, Specificity, Precision, F1-score, Visualisation &
  BIDMC Congestive Heart Failure, MIT-BIH Arrhythmia, PTB &
  Addressing Data unavailability, Privacy preservation \\
  \textbf{Herranz et al, 2025} \cite{del2025synthetic} &
  Conditional GAN &
  Kolmogorov-Smirnov test, Jensen-Shannon distance, and Pairwise correlation &
  BHI 2024 Data Challenge &
  Addressing Data unavailability, Privacy preservation \\
  \textbf{Villaizan et al, 2025} \cite{villaizan2025diffusion} &
  Conditional Diffusion model &
  Wassertstein distance, Jensen-Shannon distance, Pairwise correlation, Pearson correlation coefficient, Theil uncertainty coefficient, Distance to closest record (DCR) &
  Custom dataset &
  Data Augmentation \\
  \textbf{Kannan et al, 2025} \cite{kannan2025enhancement} &
  SMOTE, ADASYN, CTGAN, and Deep-CTGAN &
  AUC-ROC, F1-score &
  Dengue, Covid, Kidney &
  Data Augmentation, Privacy preservation \\
  \textbf{Bin Tarek et al, 2025} \cite{bin2025fairness} &
  FairSynth &
  disparate impact (DI), worst-performing true positive rate (WTPR) &
  MIMIC-III, PIC &
  Checking bias \\
  \textbf{Hao et al, 2025} \cite{hao2025llmsyn} &
  Large Language Model (LLM) &
  Accuracy, F1-score, AUC-ROC &
  MIMIC-III &
  Generate records without patient-level data \\
  \textbf{Batreddy et al, 2025} \cite{batreddy2025ehr} &
  Normalizing Flow &
  Total Variation Distance (TVD), Kullback–Leibler Divergence, Jensen–Shannon Divergence, Membership Inference Risk, Attribute Disclosure Risk, Distance to Closest Record, AUC, F1-score &
  MIMIC-IV, eICU &
  Conditional EHR trajectory generation \\
  \textbf{Kang et al, 2025} \cite{kang2025tabular} &
  GAN &
  AUC &
  Lung and Liver Cancer datasets &
  Considering inter-column relationships, Privacy preservation \\
  \textbf{Wang et al, 2025} \cite{wang2025addressing} &
  VAE &
  Area Under the Precision–Recall Curve (PRAUC), AUC-ROC, Balanced Accuracy (BA), F1-score &
  Glioma, Maternal Health, HCV, Contraceptive, Myocardial Infarction, Cardiotocography, AIDS, NHNH, SUPPORT2, EEG Eye, Diabetes, CDC Diabetes &
  Dataset balancing \\
  \textbf{Li et al, 2025} \cite{Li2025PopulationAD} &
  Diffuision model &
  Discriminative Accuracy (DA) &
  Sines, Stocks, Energy &
  Checking bias, Data Augmentation, Privacy preservation,  \\
  \textbf{Lim et al, 2025} \cite{Lim2025TSGMRA} &
  Diffusion Model, Score-based Generative Model &
  Predictive Score, Discriminative Score, Kernel Density Estimation (KDE), t-SNE &
  Stock, Energy, Air, AI4I (AI4I 2020 Predictive Maintenance dataset) &
  Data Augmentation, Privacy preservation \\
  \textbf{Fadlon et al, 2025} \cite{Fadlon2025ADM} &
  Diffusion model &
  Discriminative Score, Predictive Score, Mean Absolute Error (MAE), contextFID (Fréchet Inception Distance, t-SNE, Kernel Density Estimation (KDE) &
  Sines, Stocks, MuJoCo (physics simulation), Energy, ETTh1, ETTh2, ETTm1, ETTm2, Weather, Electricity, KDD-Cup, Traffic &
  Testing systems, Dataset balancing \\
  \textbf{Gonen et al, 2025} \cite{Gonen2025TimeSG} &
  Diffusion Model &
  Discriminative Score, Predictive Score, contextFID &
  MuJoCo, ETTm1, ETTm2, ETTh2, Sines, Weather, ILI, Saugeen River Flow, ECG200, SelfRegulationSCP1, AirQuality, StarLightCurves &
  Checking bias, Data unavailability, Privacy preservation \\
  \textbf{Suh et al, 2024} \cite{Suh2024TimeAutoDiffCA} &
  Diffusion Model &
  Discriminative Score, Temporal-correlation, Predictive metrics., Distance-to-closest-record (DCR) &
  Healthcare data, Financial market data &
  Missing data imputation \\
  \textbf{Karami et al, 2025} \cite{Karami2024SynEHRgySM} &
  Custom Method &
  Precision, Recall, Density, and Coverage (PRDC), Membership Inference Attack &
  MIMIC-III &
  Data Augmentation, Privacy preservation \\
  \textbf{Huang et al, 2025} \cite{Huang2025TimeDPLT} &
  Diffusion Model &
  Maximum Mean Discrepancy (MMD), Kullback-Leibler divergence, Marginal Distribution Difference (MDD) &
  Electricity, Solar, Wind (Energy); Traffic, Taxi, Pedestrian (Transport); Air Quality, Temperature, Rain (Nature), NN5, Fred-MD, Exchange (Economic) &
  Data Augmentation, Privacy preservation, Simulation studies \\
  \textbf{Ye et al, 2025} \cite{Ye2025NonstationaryDF} &
  Diffusion Model &
  Continuous Ranked Probability Score (CRPS), Quantile Interval Coverage Error (QICE) &
  ETTh1, ETTh2, ETTm1, ETTm2, EXG (Exchange Rate), ILI (Influenza-Like Illness), ECL (Electricity), Solar, Traffic &
  Time Series Forecasting \\
  \textbf{Chen et al, 2025} \cite{Chen2025CHIMECH} &
  Diffusion Model &
  Context-Fréchet Inception Distance (Context-FID), Correlation Matrices, Discriminative scores, Predictive scores &
  Energy, ETTh (ETTh1, ETTh2), ETTm (ETTm1, ETTm2), fMRI, MuJoCo, Sines, Stocks, Weather, Electricity, Traffic &
  Data unavailability, Privacy preservation, Time Series Forecasting \\
  \textbf{Lu et al, 2025} 
  \cite{lu2025medkrg} &
  Transformer encoder, Knowledge graph &
  Domain expert review &
  Jarvis-D, Jarvis-D2  &
  Data unavailability, Privacy preservation \\
  \textbf{Si et al, 2025} 
  \cite{Si2025TabRepTT} &
  Diffusion Model &
  Machine learning efficiency, Membership inference attacks &
  Stroke, Diabetes &
  NA \\
  \textbf{Li et al, 2025} \cite{Li2025TabTreeFormerTD} &
  Transformer &
  MLE, Shape and Trend metrics, Distance to closest record (DCR), Efficiency (Model size, Computation time) &
  Breast, Diabetes &
  Capturing correlation, Privacy preservation \\
  \textbf{Zhang et al, 2025} \cite{Zhang2025CausalDiffTabMC} &
  Diffusion Model &
  alpha-Precision, beta-Recall, CS2T scores, Distance to closest record (DCR) &
  Diabetes &
  Privacy preservation \\
  \textbf{Tian et al, 2024} \cite{tian2024reliable} &
  Diffusion Model &
  t-SNE, UMAP, Discriminative and Predictive Scores, Nearest Neighbor Adversarial Accuracy Risk (NNAA), Membership Inference Risk (MIR) &
  MIMIC-III, eICU, Stocks, Energy &
  General \\
  \textbf{Deng et al, 2025} \cite{deng2025tardiff} &
  Diffusion Model &
  AUROC, AUPRC, Influence Guidance under Class Imbalance, Complexity Analysis &
  MIMIC-III, eICU &
  General \\
  \textbf{Zhong et al, 2024} \cite{zhong2024meddiffusion} &
  Diffusion Model &
  PR‑AUC, F1, Cohen’s Kappa &
  Kidney, COPD, Amnesia, MIMIC &
  Data Augmentation \\

  \textbf{Kotelnikov et al, 2023} \cite{kotelnikov2023tabddpm} &
  Diffusion Model &
  Dimension-wise distribution, Correlations between features, mean Distance to Closest Record (DCR) &
  Adult, Abalone, Buddy, California Housing, Cardio, Churn Modeling, Default, Diabetes, Facebook Comm. Vol., Gesture Phase, Higgs Small, House 16H, Insurance, King, MiniBooNE, Wilt &
  General \\
  \textbf{Shi et al, 2024} 
  \cite{shi2024tabdiff} &
  Diffusion Model &
  Shape, Trend, alpha-Precision, beta-Recall, Detection, Distance to Closest Records (DCR) &
  Adult, Default, Shoppers, Magic, Faults, Beijing, News, Diabetes &
  General \\
  \textbf{Zhong et al, 2024} \cite{zhong2024synthesizing} &
  Diffusion Model &
  AUPR (Area under the Precision-Recall curve), F1-score, Kappa, Time Interval Prediction &
  MIMIC-III, Breast Cancer Trial &
  General \\
  \textbf{Zhang et al, 2023} \cite{zhang2023mixed} &
  Diffusion Model &
  Column-wise density estimation, Pair-wise column correlation, alpha-precision, beta-recall, Privacy protection &
  Adult, Default, Shoppers, Magic, Faults, Beijing, News &
  General 
  \\
\hline
\caption{Publications included in this review and their characteristics.}
\label{tab:my-table3}
\end{longtable}
\end{small}

\end{document}